\documentclass{raw/tADR2e}

\usepackage[hyphens]{url} % For long URL.
\usepackage{comment} % For comments/highlights.
\usepackage{enumerate}
\usepackage{algorithm} % For algorithm environments.
\usepackage{algpseudocode} % For algorithmic environments.

\DeclareMathOperator*{\argmax}{argmax}

\usepackage{color}    
\usepackage{umoline} 
\usepackage{proofread} %
\noproofreadmark 

\allowdisplaybreaks

\begin{document}

\jvol{}
\jnum{}
\jyear{}
\jmonth{}

%%%%%%%%%%%%%%%%%%%%%%%%%%%%%%%%%%%%%%%%%%%%%%%%%%%%%%%%%%%%%%%%%%%%%%%%%%%%%%%%

% First pass check by Lotfi El Hafi on July 22th, 2019.
% Second pass check by Lotfi El Hafi on November 6th, 2019.

\title{
    Autonomous Planning Based on Spatial Concepts to Tidy Up \\Home Environments with Service Robots
}

\author{
    Akira Taniguchi$^{a}$$^{\ast}$\thanks{$^\ast$Corresponding author. Email: a.taniguchi@em.ci.ritsumei.ac.jp\vspace{6pt}},
    Shota Isobe$^{a}$, Lotfi El Hafi$^{a}$, 
    Yoshinobu Hagiwara$^{a}$, and Tadahiro Taniguchi$^{a}$ \\
    \vspace{6pt}
    $^{a}${\em{College of Information Science and Engineering, Ritsumeikan University, Shiga, Japan}}\\
    \vspace{6pt}
    \received{v2.0 released October 2020} 
}

\maketitle

%%%%%%%%%%%%%%%%%%%%%%%%%%%%%%%%%%%%%%%%%%%%%%%%%%%%%%%%%%%%%%%%%%%%%%%%%%%%%%%%

% First pass check by Lotfi El Hafi on July 22th, 2019.
% Second pass check by Lotfi El Hafi on November 6th, 2019.
% Third pass check by Lotfi El Hafi on December 11th, 2019.

\begin{abstract}
    Tidy-up tasks by service robots in home environments are challenging in robotics applications because they involve various interactions with the environment.
    In particular, robots are required not only to grasp, move, and release various home objects but also to plan the order and positions for placing the objects.
    In this paper, we propose a novel planning method that can efficiently estimate the order and positions of the objects to be tidied up by learning the parameters of a probabilistic generative model.
    The model allows a robot to learn the distributions of the co-occurrence probability of the objects and places to tidy up using the multimodal sensor information collected in a tidied environment.
    Additionally, we develop an autonomous robotic system to perform the tidy-up operation.
    We evaluate the effectiveness of the proposed method by an experimental simulation that reproduces the conditions of the Tidy Up Here task of the World Robot Summit 2018 international robotics competition.
    The simulation results show that the proposed method enables the robot to successively tidy up several objects and achieves the best task score among the considered baseline tidy-up methods.

    \medskip
\end{abstract}

%%%%%%%%%%%%%%%%%%%%%%%%%%%%%%%%%%%%%%%%%%%%%%%%%%%%%%%%%%%%%%%%%%%%%%%%%%%%%%%%

% First pass check by Lotfi El Hafi on July 22th, 2019.
% Second pass check by Lotfi El Hafi on November 6th, 2019.
% Third pass check by Lotfi El Hafi on December 11th, 2019.

\section{Introduction}
\label{sec:introduction}
Service robots are expected to perform daily-life support tasks in indoor home environments, such as tidying up cluttered objects, which is one of the most common tasks aimed at supporting humans at home.
Enabling robots to tidy up is considered as one of the solutions to assist aging populations with decreasing workforce, which is particularly true in Japan.
{Additionally, {devising a plan to tidy up} various types of objects in the surrounding environment is still a challenging general problem in robotics applications.}
{
In this regard, international robotics competitions related to tidy-up tasks have recently increased to accelerate the development of solutions related to these issues.
For example, the Tidy Up My Room Challenge\footnote{{Tidy Up My Room Challenge (ICRA 2018):~{\url{https://icra2018.org/tidy-up-my-room-challenge/}}}} was held at the IEEE International Conference on Robotics and Automation (ICRA) 2018.
Additionally, the Tidy Up Here\footnote{{Tidy Up Here (WRS 2018):~\url{https://worldrobotsummit.org/en/wrc2018/service/partner_real.html}\\The same task will be held at WRS 2020:
\url{https://worldrobotsummit.org/en/wrs2020/challenge/service/partner.html}}} task was one of the tasks conducted in the Partner Robot Challenge (Real Space) at the World Robot Summit (WRS) 2018~\cite{WRS2018}.
As a result of these efforts made in recent years by the competition communities, tidy-up tasks have attracted attention in the field of robotics~\cite{Jiang2012,ref:hatori2018interactively,ref:furuta2018everyday,Rasch2019}.
}

\subsection*{{Task definition}}
\begin{figure}
    \centering
    \includegraphics[width=1.00\textwidth]{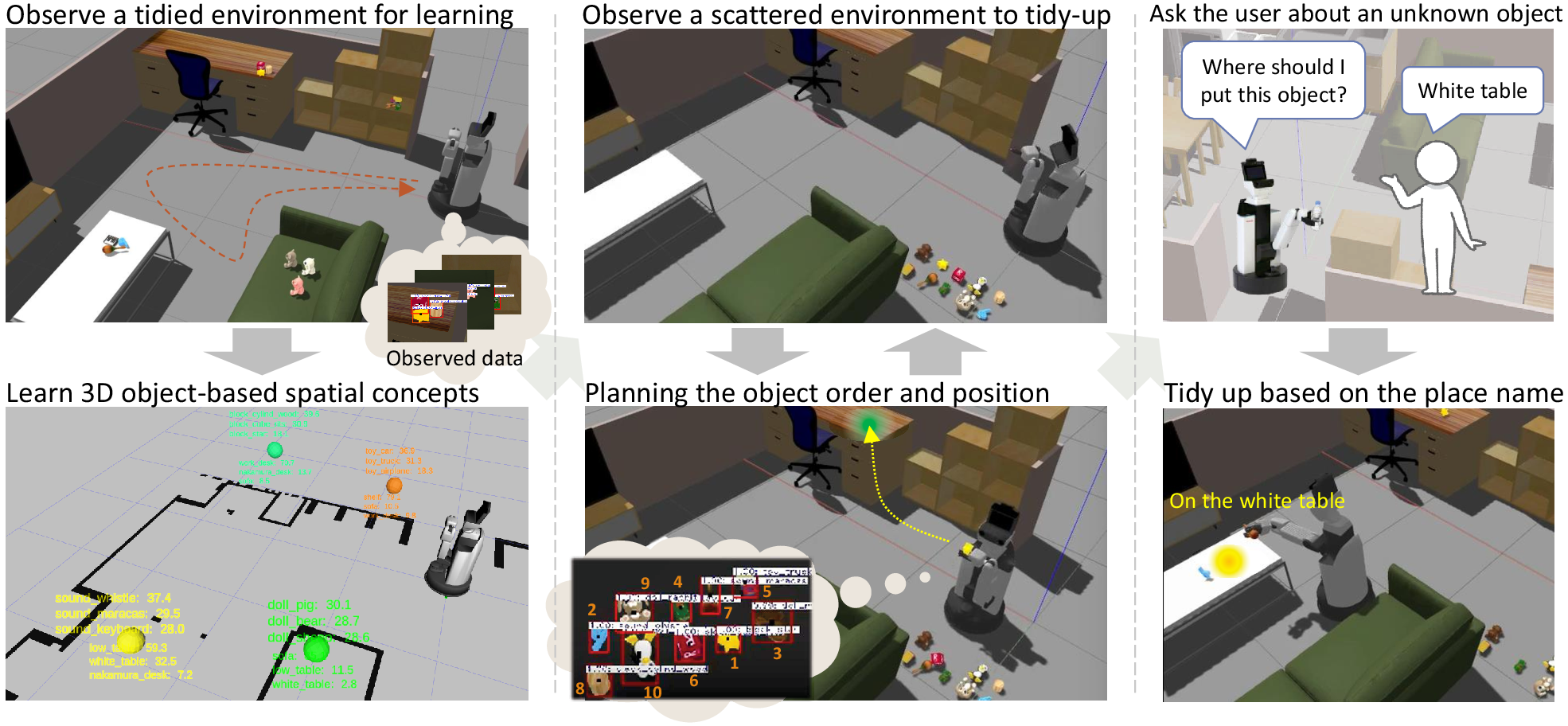}
    \caption{
        {Overview of tidy-up task in this study.
        Our approach is divided into learning and planning phases.
        (Left) In the learning phase, the robot observes the tidied environment.
        Spatial concepts are formed from observed data regarding objects and their 3D positions by multimodal learning. 
        (Middle) In the planning phase, the robot observes scattered objects in the environment, and estimates the order and positions of objects to be tidied up based on the spatial concepts formed in the learning phase.
        According to the above flow, the robot tidies up a selected object to the target position repeatedly.
        (Right) In the case of an unknown object, the robot tidies it up by asking the user about its place name.
        }
    }
    \label{fig:overview_our_study}
\end{figure}

{
We present a brief introduction of the tidy-up task considered in this study, whose overview is shown in Fig.~{\ref{fig:overview_our_study}}.
Within the scope of robotic tidy-up tasks, we focus on the planning and ordering of the arrangement of objects.
}
We reproduce the conditions of the Tidy Up Here task of the WRS 2018, the abovementioned international robotics competition, in our tidy-up evaluation experiment~{\cite{WRS2018}}.
The tidy-up task presented in this study consists of a robot moving objects located at wrong positions to their appropriate places in a room in which several objects are scattered.
In this context, wrong implies that the positions of the objects are different from their typical tidied ones.
In such a case, it is important to plan the order in which objects should be moved to conduct the tidy-up task efficiently.
For example, it is frequently inefficient to tidy up first the objects noticed immediately or randomly.
Instead, the object with the most defined tidied place, or the object which is the farthest from its appropriate position, should be tidied up first.
It is also necessary for robots to suitably ask their users about the place where to tidy up an object when it is indeterminate.

%%%%%%%%%%
%%%%%%%%%%

\subsection*{{Problem statement}}
{
To achieve the above task, the following problems need to be resolved:
(i) Adaptation to the on-site environment;
(ii) Observation uncertainty and recognition errors;
(iii) Planning the order to tidy up objects, including unknown objects; and 
(iv) Appropriate human-robot interaction (HRI) to lessen the burden on users.
}

{Service robots need to automatically recognize specific objects and learn the places where the objects are regarded as being tidied up in a given environment.
Conventional robot systems typically need to be provided labeled datasets of the objects and the places in advance to tidy up specific objects in the environment.
However, it is difficult to prepare such information by manual coding.}

Semantic mapping-based approaches are important for the spatial recognition and semantic understanding of an environment~{\cite{kostavelis2015semantic, landsiedel2017review}}.
To realize a tidy-up task, the robot must recognize the environment, estimate the objects to be tidied up while moving, and determine where to put the objects in the three-dimensional (3D) space.
It is also necessary to consider the case in which the objects to be tidied up or the categories of the objects change depending on the home environment constraints, e.g., the heights and levels of shelves.
Furthermore, it is not only necessary to deal with the differences in room configurations and object arrangements based on the home environment, but also with the grouping of the places based on the arrangements of the objects and names of specific areas.

{
Observational uncertainty and recognition errors are one of the issues that must be addressed in robotic applications.
Objects and their arrangements include uncertainty and vary depending on the home environment.
Especially, in a scattered environment, the success rate of object recognition tends to decrease.
}

{
One of our challenges is also to enable robots to autonomously tidy up without processing complex language understanding.
To reduce the burden on the user, robots need to be able to perform tidy-up tasks following simple user instructions, such as `Tidy up the kitchen.' as triggers.
This detail will be described in Section~\ref{sec:related-work:tidy-up}.
}

{
In this complicated task, attention must be paid to the design of the criteria for determining the positions and order when tidying up multiple objects to carry out the task efficiently.
In addition, the robot needs to timely ask the destination of an unknown object to the user, preferably later in the task, rather than at random timing.
}

\subsection*{{Hypothesis and assumptions}}
{
In the tidy-up task, we set the following hypothesis to solve the above problems:
\begin{enumerate}
    \item The probabilistic-based spatial concept model can deal with the uncertainty of the observations and the estimation errors, e.g., sensor noise, object misrecognitions, or self-localization errors.
    \item The existence probability of an object can represent the degree of clarity and ambiguity of the area to be tidied up for planning the object order and HRI.
\end{enumerate}
}

{
Spatial concepts refer to the knowledge of the place categories autonomously formed based on the multimodal information of the spatial experience acquired by the robot in the environment according to~\cite{ataniguchi_IROS2017,ataniguchi_IFAC2016,ref:taniguchi2016spatial,taniguchi2018unsupervised,ref:isobe2017learning,katsumata2019spcomapping}.
Therefore, we believe that the spatial concept works well not only for this particular task but also for other tasks related to place and object placement.
Especially, the spatial concept expressed probabilistically can deal with the uncertainty of the observations and the estimation errors.
}

{
In this study, we use a mobile robot with a camera, a depth sensor, and a front arm.
As shown in Fig.~{\ref{fig:overview_our_study}}, it is assumed that the robot can learn the arrangement of the objects in advance in a tidied environment.
An unknown object is defined as an object whose target place to tidy up is unknown or undefined.
The robot can detect object areas, including unknown objects, in an image.
We assume that the robot can know the initial positions and places of the cluttered objects in advance from the start of the tidy-up task.
}

\subsection*{{Our approach and main contributions}}
First, we propose a novel planning method based on the likelihood ratio maximization of the object positions in the learned model for efficiently tidying up.
Second, we introduce a probabilistic generative model (PGM) for spatial concept formation that autonomously learns the relationships between the objects and the places based on their positions in the tidied environment.
To realize the proposed method, we develop an automatic system that enables service robots to effectively tidy up.
From the perspective of system integration, a distinct feature of this study is to introduce probabilistic inference by a multimodal learning-based PGM for spatial concept formation into an integrated robotic system for tidy-up tasks.

The main contributions of this study are as follows:
\begin{enumerate}
    \item {We show that the robot can effectively use the spatial knowledge learned autonomously from sensor observations to tidy up an environment, without specifying the object positions manually.}
    \item {We show that the simultaneous estimation of objects and their tidied positions using the observation likelihood in our PGM can plan more accurately than the typical conventional methods.}
    \item {We show that the robot typically begins tidying up, in order, from the object whose target place is the most defined.} 
    \item {We show that the robot tidies up unknown objects at the end to postpone the time-consuming step of asking the user about their target places.}
\end{enumerate}

The remainder of this paper is as follows.
First, we describe the related work regarding tidy-up tasks and semantic mapping by robots in Section~{\ref{sec:related-work}}. 
Next, we describe the proposed methods in Section~{\ref{sec:proposed_method}}, and the developed system in Section~{\ref{sec:proposed_system}}.
Then, we describe the experimental conditions, evaluations, and results in Section~{\ref{sec:experiments}}.
{{In addition, we discuss the applicability of our approach and the avenues for future work in Section~\ref{sec:discussion}.}}
Finally, we give the conclusion in Section~{\ref{sec:conclusion}}.

%%%%%%%%%%%%%%%%%%%%%%%%%%%%%%%%%%%%%%%%%%%%%%%%%%%%%%%%%%%%%%%%%%%%%%%%%%%%%%%%

% First pass check by Lotfi El Hafi on July 23th, 2019.
% Second pass check by Lotfi El Hafi on November 6th, 2019.

\section{Related work}
\label{sec:related-work}

We describe the studies related to the tidy-up tasks performed by robots in home environments in Section~\ref{sec:related-work:tidy-up}.
Moreover, previous studies on spatial recognition and semantic mapping to acquire the positional relationship between objects and places are described in Section~\ref{sec:related-work:spatial}.

%%%%%%%%%%%%%%%%%%%%%%%%%%%%%%%%%%%%%%%%

% First pass check by Lotfi El Hafi on July 23th, 2019.
% Second pass check by Lotfi El Hafi on November 6th, 2019.
% Third pass check by Lotfi El Hafi on December 11th, 2019.

\subsection{Tidy-up tasks {and planning for placing} using robots}
\label{sec:related-work:tidy-up}

Some studies require time and physical efforts because the user ultimately tidies up~\cite{ref:fink2014robot,ogasawara2017stationery}.
Fink et al. evaluated whether the behavior of a robot affects the tidying behavior of children using a box-type robot to motivate them to tidy up toys on the floor~\cite{ref:fink2014robot}.
Ogasawara \& Gouko proposed a stationery holder robot to investigate its influence on the human intentions for tidying up and improve deskwork efficiency by reducing clutter~\cite{ogasawara2017stationery}.
Tidy-up tasks in home environments involve diverse shapes, weights, and places of the objects to be tidied up.
To deal with such complexity, the above studies on HRI, which motivate a user to tidy up objects, have been conducted.
In our study, we overcome the above problem of relying on a user to tidy up by proposing a planning method and a system that enables a robot to tidy up objects automatically.

{
% Autonomous learning
Additionally, it is important for robots to learn the relationships between the objects and the places only from observations, without inconveniencing the users in advance.
Abdo et al. developed a system to estimate the object arrangements in a shelf from object pair candidates based on the user preference by collaborative filtering and spectral clustering~\cite{ref:abdo2015robot}. 
Although it is possible to arrange objects based on home specificity, it is unrealistic to investigate the preferences of object pairs for each user for every destination, including places besides shelves and boxes.
Liang et al. proposed an end-user robot programming system for a robotics shelf arrangement task~\cite{Liang2018}.
In this case, a user has to teach a robot the task goals and actions by demonstration through a graphical user interface.
Hatori et al. proposed a robotic system that can manipulate an object to an intended box by interactive natural dialogue~\cite{ref:hatori2018interactively}.
However, the positions to move objects to are provided manually, and the system cannot adapt depending on the place to tidy up.
By contrast, our model learns the desired positions of the objects in a tidied scenario autonomously from sensor observation in each environment.
}

{
% Understanding utterance commands and handling unknown objects
Furthermore, a robot is required to tidy up objects autonomously without detailed linguistic instructions provided by the users in advance each time.
Alternatively, it is important for a robot to appropriately ask questions to the user about an unknown object. 
According to \cite{ref:hatori2018interactively}, when moving an object, it is necessary to issue a speech command for each object, e.g., `Put the green cup on the left end of the first level of the shelf in the kitchen.'.
In such a case, the burden on the user becomes heavy when there are multiple objects and places to tidy up in the environment.
Magassouba et al. proposed a method to understand ambiguous language instructions for carry-and-place tasks~\cite{Magassouba2018}.
In contrast, the following studies~\cite{Jiang2012,Kang2018,Rasch2019} did not deal with HRI for robotic task performance.
Jiang et al. proposed a learning approach for placing multiple objects in different placing areas in a scene~\cite{Jiang2012}.
Kang et al. presented a method enabling a robot to automatically arrange objects by task and motion planning~\cite{Kang2018}.
Rasch et al. proposed a cooperative multi-agent system for service tasks in smart environments~\cite{Rasch2019}.
In our study, it is sufficient for our method to request user interaction only for objects with undefined tidied positions, which is possible owing to the learned relationships between the objects and the places.
}

{
% Multiple object planning
Finally, when multiple objects are scattered in the environment, the robot must effectively determine the order in which the objects should be tidied up.
Furuta et al. constructed a system for local-rule-aware robot task planning to tidy up a detected object with probabilistic object location map generated by the long-term episodic memory related to the objects~\cite{ref:furuta2018everyday}.
However, the above tidy-up planning is considered only when one object is provided: it is not applicable when a plurality of general objects is observed.
Therefore, we propose a tidy-up planning method that can deal with multiple objects in an efficient order.
}

{Given these distinct aspects, we consider that our method can complement the shortcomings of the aforementioned works.}

%%%%%%%%%%%%%%%%%%%%%%%%%%%%%%%%%%%%%%%%

% First pass check by Lotfi El Hafi on July 23th, 2019.
% Second pass check by Lotfi El Hafi on November 6th, 2019.
% Third pass check by Lotfi El Hafi on December 11th, 2019.

\subsection{Semantic mapping and spatial concept formation}
\label{sec:related-work:spatial}

To realize tidy-up tasks using robots, it is necessary to deal with room configurations and object arrangements, which differ with the home environment, by spatial recognition and semantic understanding of the 3D environment~\cite{kostavelis2015semantic}.
There are studies on 3D semantic mapping~\cite{zhao2016building,ref:SemanticFusion_mccormac2017,ref:ObjectDetection_SemanticMapping_sunderhauf2017,ref:CNN-SLAM_tateno2017,ref:Multimodal_Semantic_3D_Mapping_jeong2018} that provide labels of places and objects, obtained by semantic segmentation and object detection methods, to map the information constructed from visual simultaneous localization and mapping (vSLAM)~\cite{ref:taketomi2017visual}.

The effective application of the above 3D semantic mapping methods has become an important issue in recent years as they generally require a large amount of labeled training data.
Additionally, in many of them, the pixel-level labeled data generated by the semantic segmentation of the observed images correspond directly to voxels in a metric map.
In such cases, some technical difficulties must be solved to use robots for tidy-up tasks, e.g., dealing with misrecognition without any changes, and all the voxels must be stored in a large memory.
These become even more important issues in robot systems that must perform various internal processes in real-time with limited embedded resources.

It is important to appropriately generalize and form place categories based on object positions while dealing with the uncertainty of the observations.
To solve these issues, unsupervised learning approaches for spatial concepts were utilized in studies related to autonomous place categorization by a robot~\cite{ataniguchi_IROS2017,ataniguchi_IFAC2016,ref:taniguchi2016spatial,taniguchi2018unsupervised,ref:isobe2017learning,katsumata2019spcomapping}.
Taniguchi et al. proposed nonparametric Bayesian spatial concept acquisition methods, SpCoA~\cite{ref:taniguchi2016spatial} and SpCoA++~\cite{taniguchi2018unsupervised}, which integrate self-localization and unsupervised word-segmentation from speech signals as PGMs through the latent variables of spatial concepts.
Their methods improve the accuracy of self-localization and recognition of the place names in spoken sentences.
Moreover, SpCoSLAM~\cite{ataniguchi_IROS2017} realizes online learning of spatial concepts, language models, and maps.
Katsumata et al. proposed a Markov random field-based statistical semantic mapping method, SpCoMapping~\cite{katsumata2019spcomapping}, for the segmentation of place categories in a two-dimensional (2D) grid map to determine the area to be cleaned by vacuum cleaner robots.
Such a PGM approach for the spatial concept formation was also adopted for navigational tasks in the Future Convenience Store Challenge of the WRS 2018~\cite{lotfi2018WRS}.
However, these studies are mainly focused on the place category formation and the learning of words corresponding to these places, and did not use the learned knowledge for tidying up objects.

More recently, Isobe et al. proposed a model to learn the relationship between an object and a location using the self-position of the robot, object, and word information~\cite{ref:isobe2017learning}.
The relationship between an object and a location yields the existence probability of the object and the occurrence probability of the place name in each spatial area.
This suggests the possibility to estimate the existence of an object at in a specific location based on ambiguous commands, e.g., `Tidy up this object.'
However, problems occur when the spatial concepts are learned based on the robot's self-position with object information including not only the objects at the location of interest but also far objects visible from the viewpoint of the robot.
Moreover, the 2D position distribution of the spatial concepts on the floor plane could not accommodate the heights of shelves and desks.
Furthermore, the above studies did not provide a concrete tidy-up planning method in an actual tidy-up task.
In our study, we solve the above problem by learning the spatial concepts based on the 3D position information of the objects inside a tidied environment, instead of the robot's self-position.
Additionally, we formulate a planning algorithm based on the probabilistic inference of tidied positions of the objects.

%%%%%%%%%%%%%%%%%%%%%%%%%%%%%%%%%%%%%%%%%%%%%%%%%%%%%%%%%%%%%%%%%%%%%%%%%%%%%%%%

% First pass check by Lotfi El Hafi on August 8th, 2019.
% Second pass check by Lotfi El Hafi on November 7th, 2019.

\section{Proposed tidy-up planning based on probabilistic generative model for spatial concept formation}
\label{sec:proposed_method}

We introduce the PGM for the spatial concept formation for learning the tidied positions of objects in a tidied environment.
We describe the variable definition and the generative process of the PGM for spatial concept formation in Section~\ref{sec:proposed_method:model} and the inference procedure for the parameters of the spatial concept model in Section~\ref{sec:proposed_method:learning}.
Moreover, we propose a tidy-up planning method that can efficiently estimate the order and positions of the objects to be tidied up based on the learning results of the model parameters in the PGM.
In Section~\ref{sec:proposed_method:planning_formulation_one}, 
we describe the formulation of the proposed tidy-up planning.
In Section~\ref{sec:proposed_method:planning_unknown}, we describe the method to tidy up an unknown object by human-robot speech interaction.
Algorithm~\ref{alg:planning} is the planning algorithm used for tidying up based on the proposed method, which is obtained as described in Section~\ref{sec:proposed_method:planning_formulation_one} and \ref{sec:proposed_method:planning_unknown}.

%%%%%%%%%%%%%%%%%%%%

% First pass check by Lotfi El Hafi on August 8th, 2019.
% Second pass check by Lotfi El Hafi on November 7th, 2019.

\subsection{PGM of spatial concepts for tidying up objects}
\label{sec:proposed_method:model}

\begin{figure}
    \centering
    \includegraphics[width=0.4\textwidth]{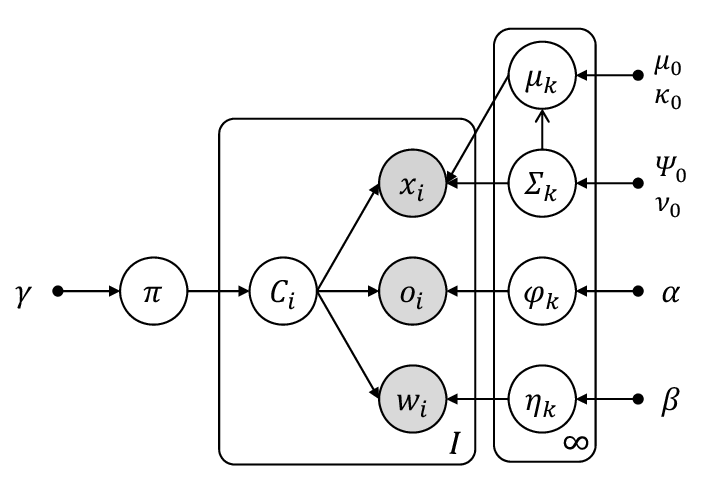}
    \caption{
        Graphical model representation of spatial concept formation for tidying up objects.
        This model is multimodal Dirichlet process mixture whose emission distributions are Gaussian and multinomial distributions.
        Gray nodes denote observed variables, and white nodes unobserved latent variables.
    }
    \label{fig:graphical_model}
\end{figure}

\begin{table}
    \tbl{
        Description of variables in generative model.
    }{
    \begin{tabular}{cl}
        \hline\noalign{\smallskip}
        \textbf{Symbol} & \textbf{Definition}\\
        \noalign{\smallskip}\hline\noalign{\smallskip}
        $x_{i}$ & Position of an object {($x_{i}=( \rm{x}_{i}, \rm{y}_{i}, \rm{z}_{i} )$)} \\
        $o_{i}$ & Detected object information {($o_{i}=( o_{i}^{1}, o_{i}^{2}, \dots , o_{i}^{{S}} )$)} \\
        $w_{i}$ & Words representing place names {corresponding to the position {$x_{i}$}} \\
         & {($w_{i}=( w_{i}^{1}, w_{i}^{2}, \cdots , w_{i}^{M} )$)} \\
        $C_{i}$ & Latent variable for the index of spatial concepts \\
        $\mu_{k}, \Sigma_{k}$   & Parameters of the Gaussian distribution for the position of object $x_{i}$ \\
        $\varphi_{k}, \eta_{k}$ & Parameters of the multinomial distribution for the observations $o_{i}$, $w_{i}$ \\
        $\pi$ & Parameter of the multinomial distribution for index $C_{i}$ of spatial concepts \\
        {$\mu_{0}, \kappa_{0}, \psi_{0}, \nu_{0}$} & Hyperparameters of the Gaussian--inverse--Wishart prior distribution \\
        {$\alpha, \beta, \gamma$} & Hyperparameters of the Dirichlet prior distribution \\
        \noalign{\smallskip}\hline\noalign{\smallskip}
        {$i$} & {Index number of training data ({$i \in \{ 1,2, \dots, I\}$})} \\
        {$k$} & {Index number of spatial concepts  ({$k \in \{ 1,2, \dots, K\}$})} \\
        {$I$} & {Total number of training data} \\
        {$K$} & {Upper limit of the number of spatial concepts} \\
        {$S$} & {Number of object classes} \\
        {$M$} & {Total number of words representing the places provided in the user's speech} \\
        \noalign{\smallskip}\hline
    \end{tabular}
    }
    \label{tab:element_of_graphical_model}
\end{table}

We introduce the PGM of spatial concepts for tidying up objects.
This model can determine the places where to tidy up in the 3D space as well as the occurrence probability of the objects and words for each place based on the positions, image, and word information related to the objects in a tidied environment.
Specifically, the model can perform place categorization in an unsupervised manner from the multimodal data obtained by a robot autonomously, i.e., it generalizes the place where an object should be tidied up from the observations regarding the objects.

Fig.~\ref{fig:graphical_model} shows the graphical model, and Table~\ref{tab:element_of_graphical_model} summarizes the description of the random variables in the model.
We describe below the generative process of the model:
\begin{eqnarray}
    \pi &\sim& {\rm GEM}\left( \gamma \right) \label{eq:generation_model_start}\\
    C_{i} &\sim& {\rm Mult} \left( \pi \right)\\
    \Sigma_{k} &\sim& \mathcal{IW} \left( \psi _{0}, \nu _{0} \right)\\
    \mu_{k} &\sim& \mathcal{N} \left( \mu _{0},  \Sigma_{k} /\kappa _{0} \right)\\
    \varphi_{k} &\sim& {\rm Dir}\left( \alpha \right)\\
    \eta_{k} &\sim& {\rm Dir}\left( \beta \right)\\
    x_{i} &\sim& \mathcal{N}\left( \mu _{C_{i}}, \Sigma _{C_{i}} \right)\\
    o_{i} &\sim& {\rm Mult}\left( \varphi _{C_{i}} \right)\\
    w_{i} &\sim& {\rm Mult}\left( \eta _{C_{i}} \right) 
    \label{eq:generation_model_end}
\end{eqnarray}
where the stick-breaking process (SBP)~\cite{sethuraman1994constructive,ref:ishwaran2001gibbs}, a type of Dirichlet process, is denoted as ${\rm GEM}(\cdot )$, the multinomial distribution as ${\rm Mult}(\cdot )$, Dirichlet distribution as ${\rm Dir}(\cdot )$, inverse--Wishart distribution as $\mathcal{IW}(\cdot )$, and multivariate Gaussian distribution as $\mathcal{N}(\cdot )$.
Refer to \cite{murphy2012machine} for the specific formulas of the above probability distributions.

In a previous method~\cite{ref:isobe2017learning}, the position distribution was represented by 2D coordinates.
In contrast, the proposed method learns the relationship between the objects and the places in the 3D space and can deal with cases where the objects to be tidied change depending on the height, e.g., on shelves and tables.
In this study, we adopt the weak-limit approximation~\cite{fox2011sticky} in the SBP.
The appropriate number of spatial concepts is probabilistically determined by learning based on the observations.

%%%%%%%%%%%%%%%%%%%%

% First pass check by Lotfi El Hafi on August 8th, 2019.
% Second pass check by Lotfi El Hafi on November 7th, 2019.

\subsection{Learning of parameters for the spatial concept model}
\label{sec:proposed_method:learning}

During the learning of spatial concepts, the robot estimates the set of all the latent variables, $\mathbf{C}=\{ C_{i} \}_{i=1}^{I}$, and the set of model parameters, $\Theta = \{ \mu_{k}, \Sigma_{k}, \varphi_{k}, \eta_{k}, \pi \}$, from the set of multimodal observations, $\mathbf{O} = \{ x_{i}, o_{i}, w_{i} \}_{i=1}^{I}$, by Gibbs sampling, which is a type of Markov Chain Monte-Carlo.
The sampling values are given by the iteration of the Gibbs sampling\footnote{We provide the details of the Gibbs sampling in Appendix~\ref{apdx:proposed_method:learning:gibbs}.} from the joint posterior distribution as follows:
\begin{eqnarray}
  \mathbf{C}, \Theta  \sim p \left(\mathbf{C}, \Theta \mid\mathbf{O}, \mathbf{h} \right)
  \label{eq:gibbs} 
\end{eqnarray}
where the set of hyperparameters is denoted as $\mathbf{h} = \{ \alpha, \beta, \gamma, \mu_{0}, \kappa_{0}, \psi_{0}, \nu_{0} \}$.

%%%%%%%%%%%%%%%%%%%%

% First pass check by Lotfi El Hafi on August 8th, 2019.
% Second pass check by Lotfi El Hafi on November 7th, 2019.

\subsection{Formulation for planning tidying of object based on spatial concepts}
\label{sec:proposed_method:planning_formulation_one}

The proposed tidy-up planning method is formulated to first select the object whose tidied place is the most defined.
In the spatial concept model learned in the tidied environment, the likelihood is the highest when each object position is tidied.
Therefore, when tidying up one object from scattered objects, the object with the highest likelihood is selected first\footnote{The formulation for the simultaneous estimation of $N$ objects and their positions for tidying up is described in Appendix~\ref{apdx:proposed_method:planning_formulation_multi}.}.
In this study, tidying up implies moving the positions of the observed objects to increase the likelihood of the spatial concept model.
A object $d^{\ast}$ and the position $x_{d}^{\ast}$ to tidy up are estimated among the detected $D$ objects, as expressed in the following equation:
\begin{eqnarray}
    d^{\ast}, x_{d}^{\ast}
    &=& \argmax _{d,x_{d}^{\prime}} L \left( \{ x_{j} \}_{j\backslash d}, x_{d}^{\prime} \right) - \underbrace{L \left( \{ x_{j} \} \right)}_{\rm const.} \label{eq:planning_greedy}
\end{eqnarray}
where the tidied position of the object $d$ is $x_{d}^{\prime}$, likelihood function on the object positions when a set of object information is given is $L \left( \{ x_{j} \} \right) = p \left( \{ x_{j} \} \mid \{o_{j}\}, \Theta \right)$, set of parameters learned by the spatial concept model is $\Theta$, and set of position information of the detected objects excluding the position information of the object $d$ is $\{ x_{j} \}_{j\backslash d}$.
{The index of the detected objects is {$j \in \{ 1,2, \dots, D \}$}.}
Additionally, we only deal with $L \left( \{ x_{j} \}_{j\backslash d}, x_{d}^{\prime} \right)$ because $L \left( \{ x_{j} \} \right)$ is constant in Equation~(\ref{eq:planning_greedy}).

The positions $x_{j}$ of each object are conditionally independent of each other.
It is assumed that the positions of the objects besides the position $x_{d}^{\prime}$ of the object to be tidied up do not change.
Therefore, $L \left( \{ x_{j} \}_{j\backslash d}, x_{d}^{\prime} \right)$ is developed as follows:
\begin{eqnarray}
    &&L \left( \{ x_{j} \}_{j\backslash d}, x_{d}^{\prime} \right) \notag\\
    &&= p \left( \{ x_{j} \}_{j\backslash d} \mid \{ o_{j} \}_{j\backslash d}, \Theta \right) p \left( x_{d}^{\prime} \mid o_{d}, \Theta \right) \notag\\
    &&\quad \propto \prod_{j}^{D \backslash d} \sum_{C_{j}} p \left( x_{j} \mid \mu_{C_{j}}, \Sigma_{C_{j}} \right) p \left( o_{j} \mid \varphi_{C_{j}} \right) p \left( C_{j} \mid \pi \right) \notag\\
    &&\qquad \times \sum_{C_{d}} p \left( x_{d}^{\prime} \mid \mu_{C_{d}}, \Sigma_{C_{d}} \right) p \left( o_{d} \mid \varphi_{C_{d}} \right) p \left( C_{d} \mid \pi \right) 
    \label{eq:likelihood_function}
\end{eqnarray}
where $D\backslash d$ indicates that the object $d$ is excluded among $D$ detected objects.

Next, we consider the issue of estimating $x_{d}^{\ast}$ based on the object $d$.
The argmax operation in Equation~(\ref{eq:planning_greedy}) is approximated as follows:
\begin{eqnarray}
    &&\argmax _{x_{d}^{\prime}} \sum_{C_{d}} p \left( x_{d}^{\prime} \mid \mu_{C_{d}}, \Sigma_{C_{d}} \right) p \left( o_{d} \mid \varphi_{C_{d}} \right) p\left( C_{d} \mid \pi \right) \notag\\
    &&\approx \sum_{C_{d}} \underbrace{\int x_{d}^{\prime} \: p\left( x_{d}^{\prime} \mid \mu_{C_{d}}, \Sigma_{C_{d}} \right) dx_{d}^{\prime}}_{\mu_{C_{d}}} p\left( o_{d} \mid \varphi_{C_{d}} \right) p\left( C_{d} \mid \pi \right).
    \label{eq:approx_select_xd_best}
\end{eqnarray}
Furthermore, the operation of the marginalization of $C_{d}$ is approximated as the value with the highest probability as follows:
\begin{eqnarray}
    C_{d}^{\ast} &=& \argmax _{C_{d}} p \left( o_{d} \mid \varphi_{C_{d}} \right) p \left( C_{d} \mid \pi \right), \label{eq:select_xd_best}\\
    x_{d}^{\ast} &=& \mu_{C_{d}^{\ast}}. \label{eq:xd_best}
\end{eqnarray}

Moreover, because $x_{d}^{\ast}$ in Equation~(\ref{eq:planning_greedy}) is obtained by Equation~(\ref{eq:xd_best}), the object $d$ is obtained when $L \left( \{ x_{j} \}_{j\backslash d}, x_{d}^{\prime} = \mu_{C_{d}^{\ast}} \right)$ becomes the maximum as follows:
\begin{eqnarray}
    &&L \left( \{ x_{j} \}_{j\backslash d}, x_{d}^{\prime} = \mu_{C_{d}^{\ast}} \right) \notag\\
    &&= p \left( \{ x_{j} \}_{j\backslash d} \mid \{ o_{j} \}_{j\backslash d}, \Theta \right) p \left( x_{d}^{\prime} = \mu_{C_{d}^{\ast}} \mid o_{d}, \Theta \right) \notag\\
    &&\quad\propto \prod_{j}^{D\backslash d} \sum_{C_{j}} p \left( x_{j} \mid \mu_{C_{j}}, \Sigma_{C_{j}} \right) p \left( o_{j} \mid \varphi_{C_{j}} \right) p \left( C_{j} \mid \pi \right) \notag\\
    &&\qquad \times \sum_{C_{d}} p \left( x_{d}^{\prime} = \mu_{C_{d}^{\ast}} \mid \mu_{C_{d}}, \Sigma_{C_{d}} \right) p \left( o_{d} \mid \varphi_{C_{d}} \right) p \left( C_{d} \mid \pi \right). 
    \label{eq:likelihood_function_kai}
\end{eqnarray}

Additionally, we consider the case of selecting the $n$-th object to be tidied up when $n-1$ objects have been tidied already.
We introduce the likelihood ratio between the likelihoods before and after tidying up concerning the detected object $d$ when tidying up the $n$-th object as follows:
\begin{eqnarray}
    R_{n}({x_d, x_{d}^{\prime}}) 
    &=& \underbrace{L \left( \{ x_{j} \}_{j\backslash d}, x_{d}^{\prime} \right)}_{\text{Likelihood after tidying up}}/\underbrace{L \left( \{ x_{j} \} \right)}_{\text{Likelihood before tidying up}} \notag\\
    &=& \cfrac{\sum_{C_{d}} p \left( x_{d}^{\prime} \mid \mu_{C_{d}}, \Sigma_{C_{d}} \right) p \left( o_{d} \mid \varphi_{C_{d}} \right) p \left( C_{d} \mid \pi \right)}{\sum_{C_{d}} p \left( x_{d}^{ } \mid \mu_{C_{d}}, \Sigma_{C_{d}} \right) p \left( o_{d} \mid \varphi_{C_{d}} \right) p \left( C_{d} \mid \pi \right)}.
    \label{eq:ratio_probability}
\end{eqnarray}
Note that $1 \le n$.

Finally, solving Equation~(\ref{eq:planning_greedy}) is equivalent to estimating the object $d$ and the position $x_{d}^{\prime}$ with the maximum likelihood ratio as follows:
\begin{eqnarray}
    \argmax _{d,x_{d}^{\prime}} L \left( \{ x_{j} \}_{j\backslash d}, x_{d}^{\prime} \right) 
    = \argmax _{d,x_{d}^{\prime}} R_{n}({x_d, x_{d}^{\prime}}).
    \label{eq:ratio_probability_kai}
\end{eqnarray}

%%%%%%%%%%%%%%%%%%%%

% First pass check by Lotfi El Hafi on August 8th, 2019.
% Second pass check by Lotfi El Hafi on November 7th, 2019.

\subsection{How to tidy up unknown objects by speech interaction with user}
\label{sec:proposed_method:planning_unknown}

When an object is determined as being unknown in Stage 2, the robot can ask the user the target place to tidy up, e.g., `Where should I put this object?'.
Additionally, the user can provide the robot a word $w_{d}$ representing the correct place when asked.
The robot then determines the object $d$ with the probability value $P_{d}$ below the threshold $\lambda$ as the object whose target place is unknown by finding the maximum of Equation~(\ref{eq:select_xd_best}) as follows:
\begin{eqnarray}
    P_{d} &=& \max_{C_{d}} p \left( o_{d} \mid \varphi_{C_{d}} \right) p \left( C_{d} \mid \pi \right), 
    \label{eq:select_xd_best_max}
    \\
    C_{d}^{\ast} &=& \argmax _{C_{d}} \left\{ \begin{array}{lll}
    p \left( o_{d} \mid \varphi_{C_{d}} \right) p \left( C_{d} \mid \pi \right) & \quad & (\lambda \le P_{d})\\
    p \left( w_{d} \mid \eta_{C_{d}} \right) p \left( C_{d} \mid \pi \right) \label{eq:select_cg_unknown} & \quad & (\lambda > P_{d})
    \end{array}. \right.
\end{eqnarray}
This means that the tidied position of the object $d$ is ambiguous if the existence probability of the object $d$ is extremely low at each tidied position.
Finally, the robot updates the tidied position of an object that is determined to be an unknown object from the given words using Equations~(\ref{eq:select_cg_unknown}) and (\ref{eq:xd_best}).

%%%%%%%%%%%%%%%%%%%%%%%%%%%%%%%%%%%%%%%%
\begin{algorithm}
    \caption{
        Planning algorithm.
    } 
    \label{alg:planning}
    \begin{algorithmic}[1]
        \State{Collect object observations $\{ o_{j} \}$ by object detection from image data}
        \For{$n=1$ to $N$}  \Comment{Number of objects to tidy up}
            \For{$d=1$ to $D$}  \Comment{Number of detected objects} 
                \State{Estimate $C_{d}^{\ast}$} \quad \Comment{Eq.~(\ref{eq:select_xd_best})}
                \State{Decide $x_{d}^{\prime}$ as $\mu_{C_{d}^{\ast}}$} \quad \Comment{Eq.~(\ref{eq:xd_best})}
                \State{Calculate likelihood ratio $R_n(x_{d},x_{d}^{\prime})$} 
                \Comment{Eq.~(\ref{eq:ratio_probability})}
            \EndFor
            \State{Select $d^{\ast}$-th object and $x_{d}^{\ast}$} \quad \Comment{Eq.~(\ref{eq:ratio_probability_kai})}
            \State{Calculate probability value $P_{d^{\ast}}$} \quad \Comment{Eq.~(\ref{eq:select_xd_best_max})}
            \If{$\lambda > P_{d^{\ast}}$} \Comment{Judge whether the object is unknown}
                \State{Observe word information $w_{d}$ from the user}
                \State{Update $C_{d}^{\ast}$} \quad \Comment{Eq.~(\ref{eq:select_cg_unknown})}
                \State{Update $x_{d}^{\ast}$} \quad \Comment{Eq.~(\ref{eq:xd_best})}
            \EndIf
            \State{Tidy up selected $d^{\ast}$-th object to estimated position $x_{d}^{\ast}$} \Comment{Section~\ref{sec:proposed_system:planning_phase}}
        \EndFor
    \end{algorithmic}
\end{algorithm}
%%%%%%%%%%%%%%%%%%%%

%%%%%%%%%%%%%%%%%%%%%%%%%%%%%%%%%%%%%%%%%%%%%%%%%%%%%%%%%%%%%%%%%%%%%%%%%%%%%%%%

% First pass check by Lotfi El Hafi on August 9th, 2019.
% Second pass check by Lotfi El Hafi on November 8th, 2019.

\section{Autonomous robotic system for tidying up multiple objects}
\label{sec:proposed_system}

\begin{figure}[tb]
    \centering
    \includegraphics[width=0.9\textwidth]{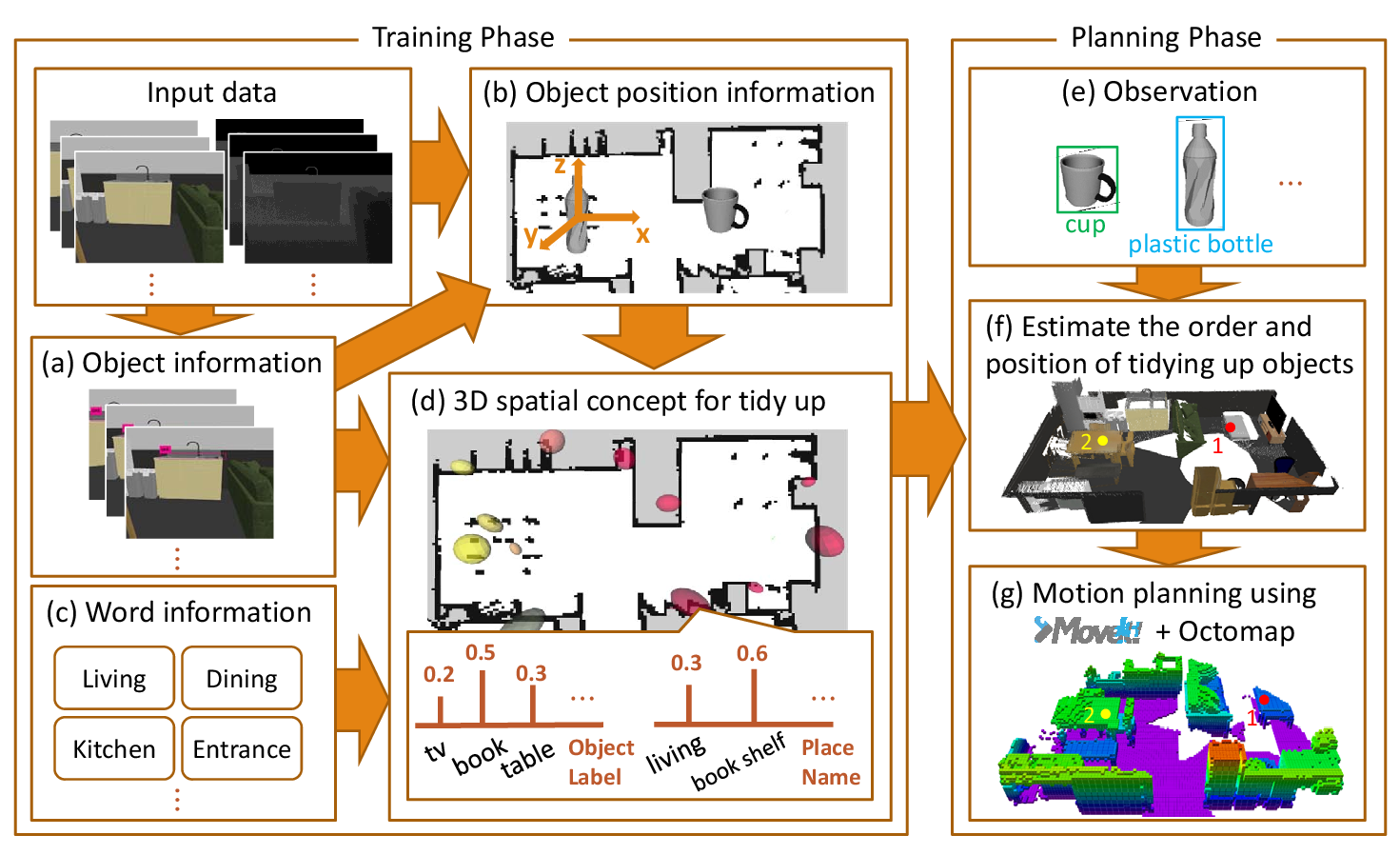}
    \caption{
        Overview diagram of the autonomous robotic system for tidying up objects.
        The system includes two phases: a training phase (left), and a planning phase (right).
    }
    \label{fig:overview_proposed_system}
\end{figure}

We develop an autonomous robotic system that can recognize and tidy up an environment automatically.
Our system is implemented with robot operating system (ROS)~\cite{ref:quigley2009ros} middleware.
An overview of the proposed autonomous robotic system is shown in Fig.~\ref{fig:overview_proposed_system}.
The training phase is described in Section~\ref{sec:proposed_system:training-phase}, and the planning phase in Section~\ref{sec:proposed_system:planning_phase}.

%%%%%%%%%%%%%%%%%%%%%%%%%%%%%%%%%%%%%%%%

% First pass check by Lotfi El Hafi on August 8th, 2019.
% Second pass check by Lotfi El Hafi on November 7th, 2019.

\subsection{Training phase}
\label{sec:proposed_system:training-phase}

The training phase consists of collecting the data and learning the spatial concepts, as shown in Fig.~\ref{fig:overview_proposed_system} (a) -- (d).

We describe the process of data collection in Fig.~\ref{fig:overview_proposed_system} (a) -- (c) as follows.
First, the robot obtains multimodal data related to the objects, i.e. positions, images, and words, in the tidied environment.
The input data in Fig.~\ref{fig:overview_proposed_system} are RGB and depth images, which are acquired by an RGB-D camera mounted on the robot. 
The robot generates a 2D occupancy grid map of the environment beforehand by Hector SLAM~\cite{ref:kohlbrecher2011flexible}, and performs self-localization by adaptive Monte-Carlo localization (AMCL)~\cite{ref:thrun2005probabilistic}.

In particular, the robot moves around the environment and acquires the following data:
\begin{itemize}
    \item[(a)] The object information, which is a one-hot vector representation of the object class corresponds to each bounding box obtained from an RGB image using a CNN-based object detection method~\cite{ref:redmon2018yolov3}.
    \item[(b)] The position information of the objects, which are 3D coordinate points of each detected object obtained with depth information from the RGB-D camera.
    Here, we use the average of the depth values of the area within 10\% of the detected bounding box center.
    \item[(c)] Word information that is a bag-of-words (BoW) representation of a place name.
    We assume that the user's speech gives the place name as one or a few isolated words representing the tidied position of the object.
\end{itemize}

In Fig.~\ref{fig:overview_proposed_system} (d), in the training phase, the spatial concept model learns the occurrence probability of the objects and words in each place based on the multimodal sensor information collected by the robot in a tidied environment, as described in Section~\ref{sec:proposed_method:learning}.

%%%%%%%%%%%%%%%%%%%%%%%%%%%%%%%%%%%%%%%%

% First pass check by Lotfi El Hafi on August 8th, 2019.
% Second pass check by Lotfi El Hafi on November 7th, 2019.

\subsection{Planning phase}
\label{sec:proposed_system:planning_phase}

In the planning phase, the robot tidies up the detected objects using the learned spatial concepts, as shown in Fig.~\ref{fig:overview_proposed_system} (e) -- (g).

First, the robot detects cluttered objects while moving.
The object detection method in the planning phase, Fig.~\ref{fig:overview_proposed_system}~(e), is the same as in the training phase, Fig.~\ref{fig:overview_proposed_system}~(a).

Next, if multiple scattered objects are detected, the robot decides on an object and a place to be tidied up, as shown in Fig.~\ref{fig:overview_proposed_system}~(f).
The robot estimates simultaneously the order and positions of the objects to tidy up from the multiple objects it observed in the cluttered environment.
It uses the parameters learned by the spatial concept model, as described in Section~\ref{sec:proposed_method:planning_formulation_one}. 
Moreover, the robot can determine if the tidied position of the object is unknown from Equation~(\ref{eq:select_cg_unknown}).
If unknown, the robot can ask where to move the object, as described in Section~\ref{sec:proposed_method:planning_unknown}.

{
Finally, the robot performs motion planning, as shown in Fig.~\ref{fig:overview_proposed_system}~(g).
After both estimating the tidied positions and planning the object order by Equation~(\ref{eq:ratio_probability_kai}), the robot is required to accurately manipulate the objects to move them.
The developed system uses the MoveIt!~\cite{ref:sucan2013moveit} framework for the motion planning of the robot arm when grasping an object.
Furthermore, the robot moves to the appropriate search positions while performing self-localization based on its map and observations.
}

When the robot finishes tidying an object, the state of the other objects may have changed owing to external factors.
To deal with such scenarios, it is possible to sequentially plan the tidy-up task by redoing the object detection after tidying each object.

%%%%%%%%%%%%%%%%%%%%%%%%%%%%%%%%%%%%%%%%%%%%%%%%%%%%%%%%%%%%%%%%%%%%%%%%%%%%%%%%

% First pass check by Lotfi El Hafi on August 9th, 2019.
% Second pass check by Lotfi El Hafi on November 8th, 2019.

\section{Experiments}
\label{sec:experiments}

We perform experiments based on the conditions and evaluation of Stages 1 and 2 of the Tidy Up Here task in WRS. 
By the quantitative evaluation of the achievement of the tidy-up task, we aim to show the validity and viability of the proposed planning method for tidying up objects.
The success of the tidy-up task is measured based on the same score criteria as the Tidy Up Here task.
Furthermore, in Stage 2, we conduct experiments not only exclusively with known objects that have already been learned (Stage~2-1) but also including objects with unknown target positions (Stage~2-2).
Herewith, we show that the robot can adequately tidy up by asking the user for the desired target place for the objects without known tidied positions.

%%%%%%%%%%%%%%%%%%%%%%%%%%%%%%%%%%%%%%%%

% First pass check by Lotfi El Hafi on August 9th, 2019.
% Second pass check by Lotfi El Hafi on November 8th, 2019.

\subsection{Condition and environmental setting}
\label{sec:experiments:condition}

\begin{figure}
    \centering
    \includegraphics[width=0.5\textwidth]{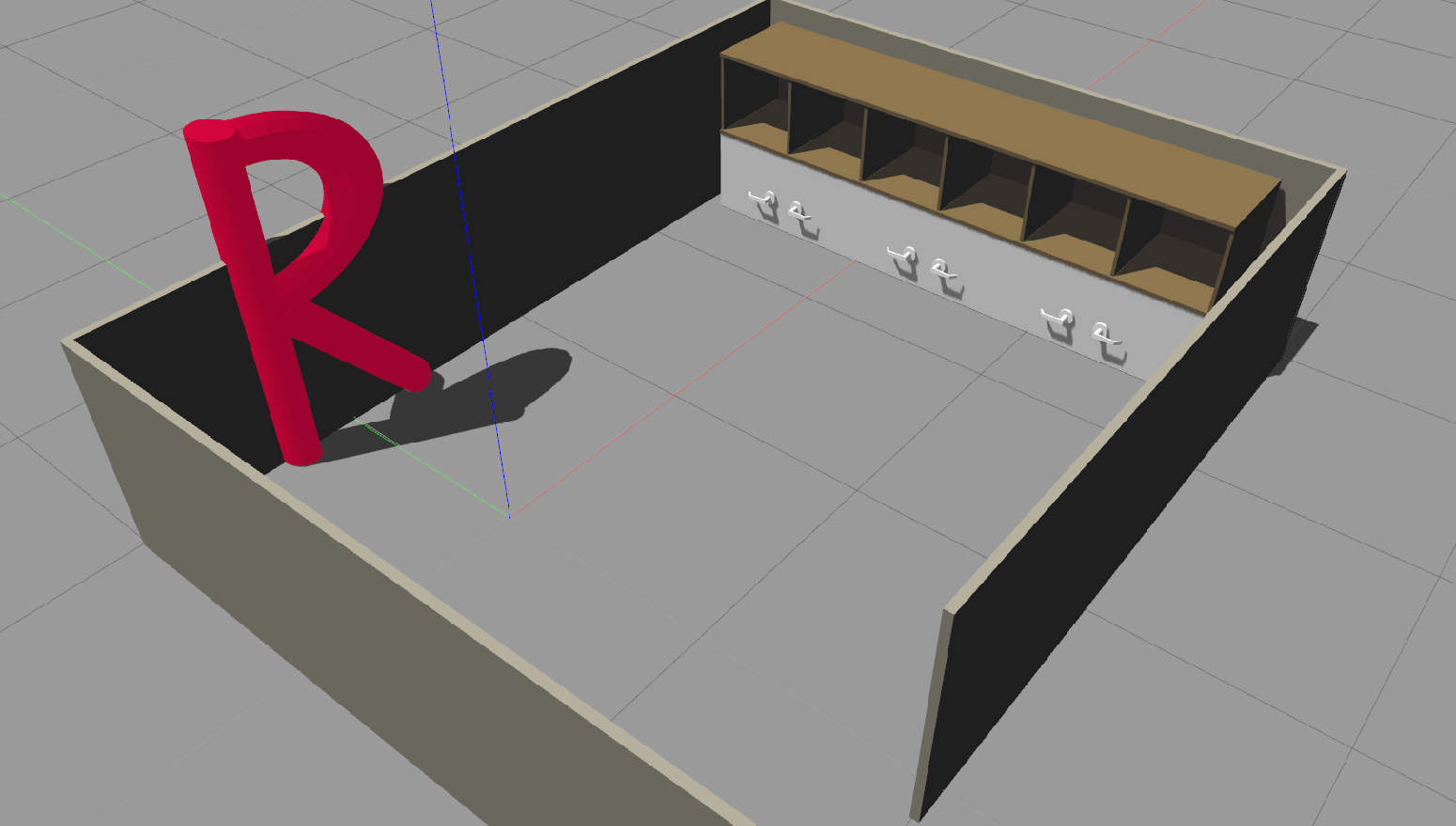}
    \caption{
        Simulator environment used for Stage~1: children's room with toy boxes.
    }
    \label{fig:env_stage1}
\end{figure}

\begin{figure}
    \centering
    \includegraphics[width=0.5\textwidth]{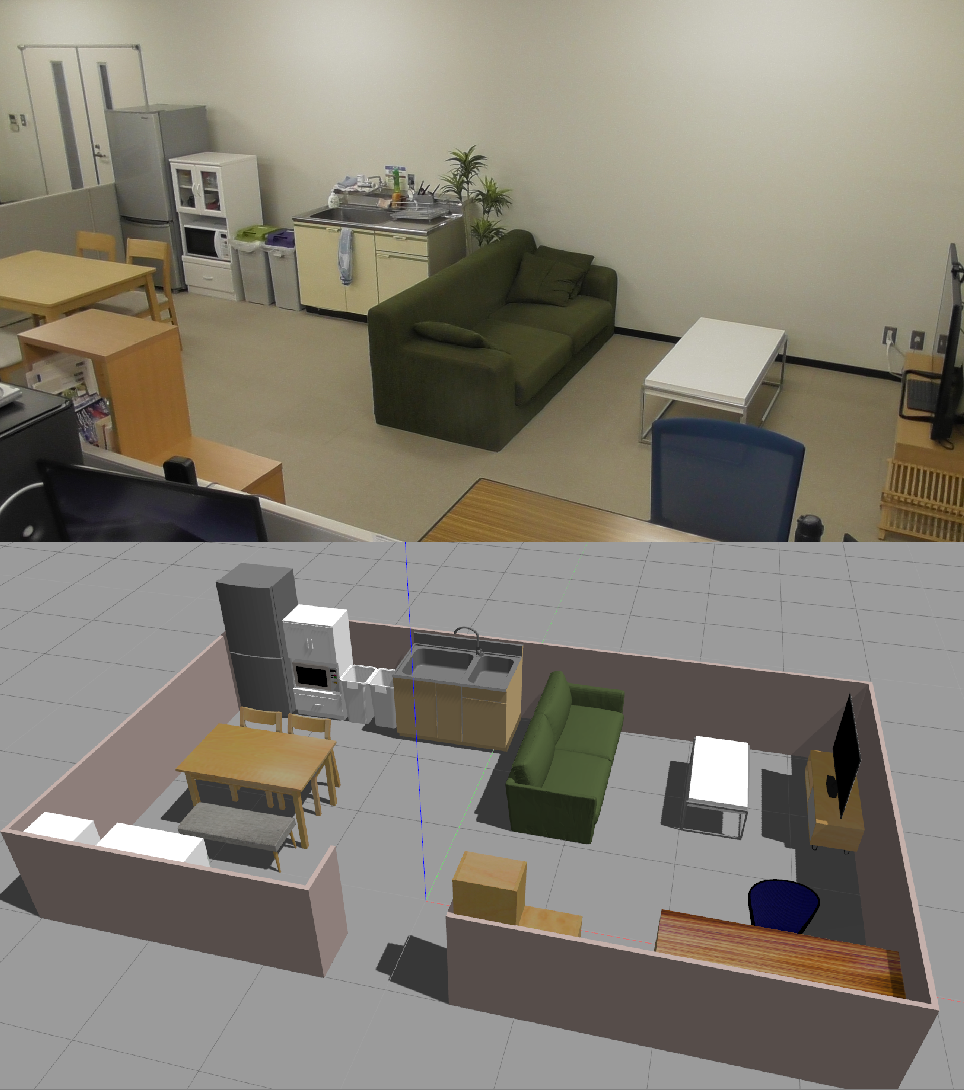}
    \caption{
        Top: real home environment replica.
        Bottom: simulator environment used for Stage~2: fully furnished home environment.
    }
    \label{fig:env_stage2}
\end{figure}

In this experiment, we use a simulator environment with a virtual Toyota human support robot (HSR)\footnote{Toyota Human Support Robot~(HSR): \url{http://www.toyota-global.com/innovation/partner_robot/robot/#link02}\\
The HSR has been adopted as a standard platform for RoboCup@Home~\cite{Iocchi2015robocup} and WRS.}~\cite{HSR2019}.
The simulations are performed on a laptop computer with the following specifications: Intel Corei7-7820HK CPU, 32 GB of DDR4 memory, Nvidia GeForce GTX 1080 GPU.
The WRS rulebook defines these specifications.
The robot is operated by the ROS Kinetic Kame running on Ubuntu 16.04 LTS.
The simulator environments constructed using Gazebo~\cite{ref:koenig2004design} are shown in Figs.~\ref{fig:env_stage1} and \ref{fig:env_stage2}.
The objects are placed at random initial positions within a specific area of the environment for each trial.

%%%%%%%%%%%%%%%%%%%%

% First pass check by Lotfi El Hafi on August 9th, 2019.
% Second pass check by Lotfi El Hafi on November 8th, 2019.

\subsubsection{Stage~1}
\label{sec:experiments:condition:stage1}

\begin{table}
    \tbl{
        3D models and names of known objects (Stages~1 and 2).
        All are known and do not deform.
    }{
    
    \begin{tabular}{lp{60mm}p{60mm}}
        & \begin{center}
            \includegraphics[height=0.08\textwidth]{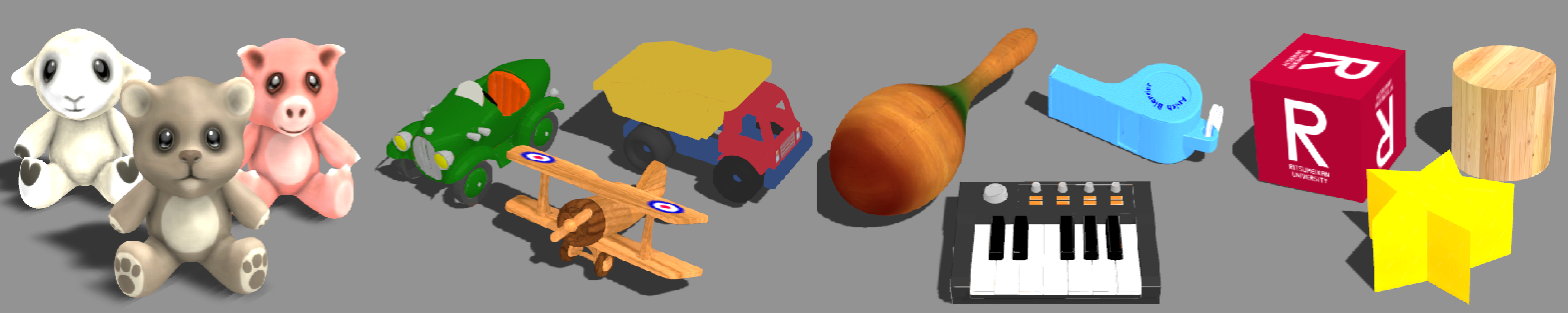}
          \end{center}\\
        \hline\noalign{\smallskip}
        \textbf{Category} & Plush doll & Toy car\\
        \noalign{\smallskip}\hline\noalign{\smallskip}
        & Bear plush toy ({\tt doll\_bear}) & Green toy car ({\tt toy\_car})\\ 
        \textbf{Object} & Sheep plush toy ({\tt doll\_sheep}) & Truck toy car ({\tt toy\_truck})\\ 
        & Pig plush toy ({\tt doll\_pig}) & Toy airplane ({\tt toy\_airplane})\\
        \noalign{\smallskip}\hline\noalign{\smallskip}
        \noalign{\smallskip}\hline\noalign{\smallskip}
        \textbf{Category} & Sound toy & Block\\
        \noalign{\smallskip}\hline\noalign{\smallskip}
        & Whistle sound toy ({\tt sound\_whistle}) & Cube block ({\tt block\_cube\_rits})\\
        \textbf{Object} & Shaker sound toy ({\tt sound\_maracas}) & Star-shaped block ({\tt block\_star})\\
        & Piano sound toy ({\tt sound\_keyboard}) & Cylinder block ({\tt block\_cylind\_wood})\\
        \noalign{\smallskip}\hline
    \end{tabular}
    }
    \label{tab:experiment_object_list}
\end{table}

\begin{table}
    \tbl{
        3D models and names of objects with unknown tidied positions (Stage~2-2).
    }{
    \begin{tabular}{lp{60mm}}
        & \begin{center}
             \includegraphics[height=0.08\textwidth]{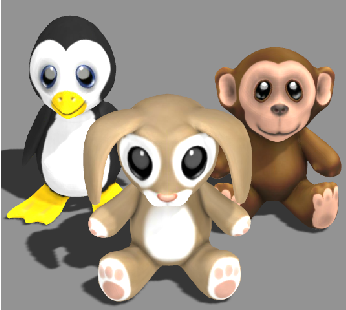}
          \end{center}\\
        \hline\noalign{\smallskip}
        \textbf{Category} & Plush doll\\
        \noalign{\smallskip}\hline\noalign{\smallskip}
        & Penguin plush toy ({\tt doll\_penguin})\\ 
        \textbf{Object} & Monkey plush toy ({\tt doll\_monkey})\\ 
        & Rabbit plush toy ({\tt doll\_rabbit})\\
        \noalign{\smallskip}\hline
    \end{tabular}
    }
    \label{tab:experiment2_unknown_objects}
\end{table}

In the environment of Stage 1 (Fig.~\ref{fig:env_stage1}), the toy boxes are lined up at the back of the children room, and the cluttered objects must be moved into them.
Table~\ref{tab:experiment_object_list} shows the objects used in Stage 1.

The number of training data is 227.
The hyperparameters of the spatial concept model are $\alpha=0.5$, $\beta=10$, $\gamma=15$, $\mu_0=\left( 2.719, -0.394, 0.655 \right)$, $\kappa_0=0.1$, $\psi_0={\rm diag} \left( 0.01, 0.01, 0.01 \right)$, and $\nu_0=1000$.
$\mu_0$ is set as the average positions of the observed objects.
The number of Gibbs sampling iterations is 100.
In the planning phase, 10 objects from the 12 objects of Table~\ref{tab:experiment_object_list} are selected for each trial.
Except the number of objects, all the other conditions remain identical to those in the Tidy Up Here task.

%%%%%%%%%%%%%%%%%%%%

% First pass check by Lotfi El Hafi on August 9th, 2019.
% Second pass check by Lotfi El Hafi on November 8th, 2019.

\subsubsection{Stage~2}
\label{sec:experiments:condition:stage2}

In the environment of Stage 2 (Fig.~\ref{fig:env_stage2}), there are multiple target places, e.g., desks and shelves, for tidying up the cluttered objects.
In Stage 2, we perform two experiments: Stage~2-1 using the same objects as in Stage 1, and Stage 2-2 in which objects from the plush doll category used in Stage 1 are replaced by other objects whose desired tidied positions are unknown.
Table~\ref{tab:experiment2_unknown_objects} shows unknown objects.

In Stage 2-2, the word information representing the place names to be tidied is given by the user.
The word list is \{{\tt shelf}, {\tt work\_desk}, {\tt nakamura\_desk}, {\tt white\_table}, {\tt low\_table}, {\tt sofa}\}.
The number of training data is 225.
Among these, the number of word information is 11, i.e. 5\% of the data.
In Stage 2-1, the hyperparameters of the spatial concept model are $\alpha=0.5$, $\beta=10$, $\gamma=10$, $\mu_0 = \left( 1.611, 0.841, 0.628 \right)$, $\kappa_0=0.1$, $\psi_0={\rm diag} \left( 0.01, 0.01, 0.01 \right)$, and $\nu_0=1000$.
In Stage 2-2, the hyperparameters of the spatial concept model are $\alpha=0.3$, $\beta=0.3$, $\gamma=10$, $\mu_0 = \left( 1.611, 0.841, 0.628 \right)$, $\kappa_0=0.1$, $\psi_0={\rm diag} \left( 0.01, 0.01, 0.01 \right)$, and $\nu_0=1000$.
$\mu_0$ is set as the average positions of the observed objects.
The number of Gibbs sampling iterations is 100.
The threshold for identifying an unknown object is $\lambda=0.003$.

%%%%%%%%%%%%%%%%%%%%%%%%%%%%%%%%%%%%%%%%

% First pass check by Lotfi El Hafi on August 9th, 2019.
% Second pass check by Lotfi El Hafi on November 8th, 2019.

\subsection{Comparison methods}
\label{sec:experiments:comparison}

We used the following three comparison methods:
\begin{enumerate}[(A)]
    \item \textbf{Proposed method}.
    \item \textbf{Baseline method 1}:
    Selection criteria for tidying an object: nearest, tidying position candidates: database.
    The nearest object is selected from the robot position as an object to be tidied up.
    The position of the object in the list of the training data matching the selected object is determined as the desired tidy-up position.
    This method is assumed as one of the typical conventional methods used even in actual competitions.
    Actually, a similar method was adopted by our team during WRS 2018.
    \item \textbf{Baseline method 2}:
    Selection criteria for tidying an object: random, tidying position candidates: random.
    An object to be tidied up and its position were selected from the list of the training data randomly.
\end{enumerate}

%%%%%%%%%%%%%%%%%%%%%%%%%%%%%%%%%%%%%%%%

% First pass check by Lotfi El Hafi on August 9th, 2019.
% Second pass check by Lotfi El Hafi on November 8th, 2019.

\subsection{Evaluation metrics}
\label{sec:experiments:evaluation}

We measure the level of achievement of tidying up with the score criteria of the Tidy Up Here task. 
{Table~\ref{tab:tidyup_score_modification} is the score table used for evaluation of the tidy-up task.} 
{In addition, we evaluate the accuracy rate to tidy up an object to the correct place, and the order in which unknown objects are tidied up in Stage 2-2.}

{
The log-likelihood has a high value if the robot can bring the object to the correct place, starting from the object with the most defined destination place.
In all the comparison methods, the log-likelihoods are obtained from Equation (\ref{eq:likelihood_function}) for all the objects in the environment using the learning result of the spatial concepts shown in Section~\ref{sub:experiment_result:training}.
}

When evaluating the tidy-up task, we execute the whole planning system including the sub-processes developed for the object grasping and release, as described in Section~\ref{sec:proposed_system}.
Additionally, we separately evaluate the tidy-up planning using the spatial concept model, which is the main contribution of this study, while omitting object manipulation.

\begin{table}
    \tbl{{
        Score table for evaluation of the tidy-up task used in this study.
        The scores are denoted as {\tt points}~{$\times$}~{\tt number of objects}.
    }}
    {{{
    \begin{tabular}{lp{100mm}r}
        \hline\noalign{\smallskip}
        &\textbf{Performance (Stage 1)} & \textbf{Score}\\
        \noalign{\smallskip}\hline\noalign{\smallskip}
        i. &Tidy up an object into the toy storage & $3\times10$\\ 
        ii. &Tidy up an object into the correct box within the toy storage & $2\times10$\\ 
        \noalign{\smallskip}\hline\noalign{\smallskip}
        \noalign{\smallskip}\hline\noalign{\smallskip}
        &\textbf{Performance  (Stage 2-1 and 2-2)} & \textbf{Score}\\
        \noalign{\smallskip}\hline\noalign{\smallskip}
        iii. &Tidy up an object to the correct place & $5\times10$\\
        \noalign{\smallskip}\hline\noalign{\smallskip}
        \noalign{\smallskip}\hline\noalign{\smallskip}
        &\textbf{Performance  (Stage 2-2)} & \textbf{Score}\\
        \noalign{\smallskip}\hline\noalign{\smallskip}
        iv. &Tidy up an unknown undeformable object & $3\times3$\\
        \noalign{\smallskip}\hline
    \end{tabular}
    }}
    }
    \label{tab:tidyup_score_modification}
\end{table}

%%%%%%%%%%%%%%%%%%%%%%%%%%%%%%%%%%%%%%%%

% First pass check by Lotfi El Hafi on August 9th, 2019.
% Second pass check by Lotfi El Hafi on November 8th, 2019.

\subsection{Results}
\label{sub:experiment_result}

We describe the experimental results\footnote{We describe the pre-training for object detection and the pre-evaluation of objects to tidy up in Appendices~\ref{apdx:experiments:pre-train} and \ref{apdx:experiments:pre-evaluation}.} of the training phase in Section~\ref{sub:experiment_result:training} and the planning phase in Section~\ref{sub:experiment_result:planning}, respectively.

%%%%%%%%%%%%%%%%%%%%

% First pass check by Lotfi El Hafi on August 9th, 2019.
% Second pass check by Lotfi El Hafi on November 8th, 2019.

\subsubsection{Result of spatial concept formation in the training phase}
\label{sub:experiment_result:training}

\begin{figure}
    \centering
    \includegraphics[width=0.6\textwidth]{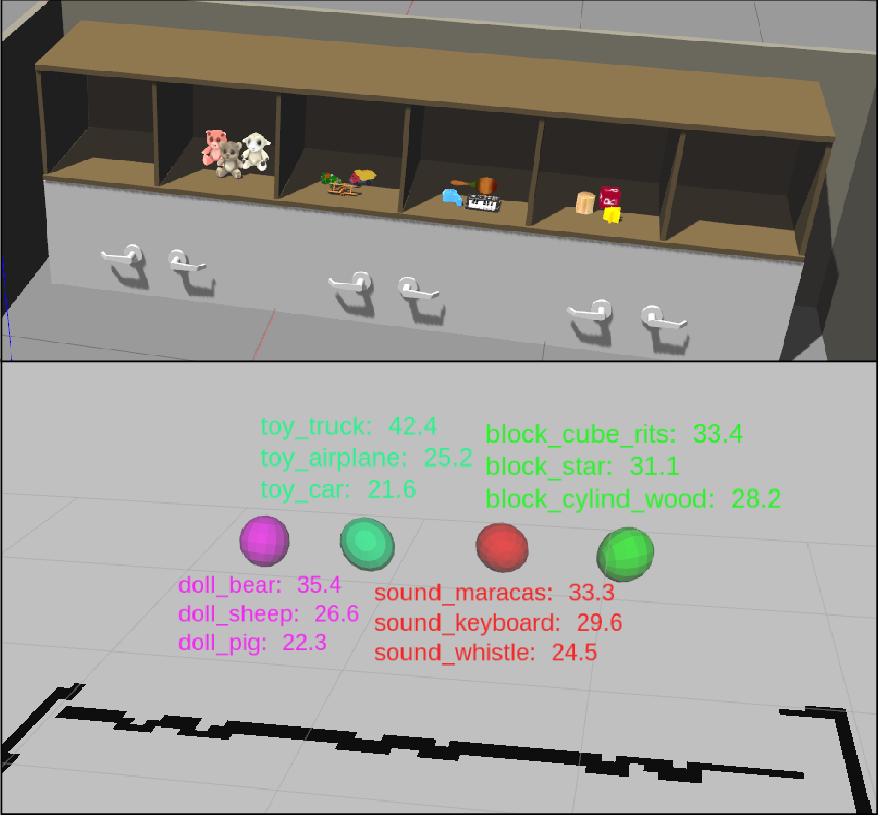}
    \caption{
        Top: object placement in tidied environment.
        Bottom: learning result of spatial concepts (Stage 1).
        Estimated distributions representing object positions are drawn as ellipsoids.
        Additionally, the three best object classes and probabilities for each spatial concept are displayed.
        Best implies highest probability value of probability distribution for object class.
        Colors are determined randomly for each index of a spatial concept.
    }
    \label{fig:spacoty_stage1_result}
\end{figure}

\begin{figure}
    \centering
    \includegraphics[width=0.6\textwidth]{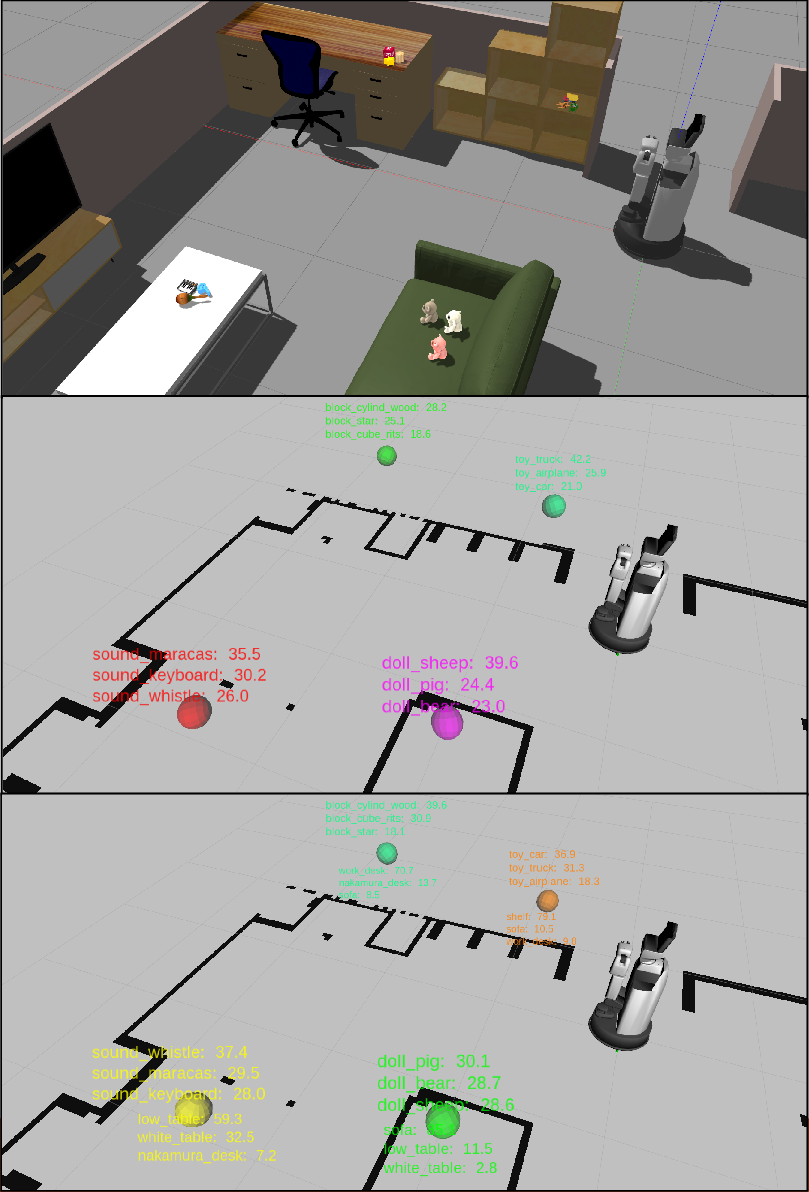}
    \caption{
        Top: object placement in tidied environment.
        Middle: learning result of spatial concepts (Stage 2-1).
        Bottom: learning result of spatial concepts (Stage 2-2).
        The three best words and probabilities for each spatial concept are displayed.
    }
    \label{fig:spacoty_stage2_result}
\end{figure}

Figs.~\ref{fig:spacoty_stage1_result} and \ref{fig:spacoty_stage2_result} show the results of the spatial concepts learned in the environments of Stages 1 and 2, respectively.
These results qualitatively show that the object classes learned at each place coincide with the objects present in the appropriately tidied environment.

%%%%%%%%%%%%%%%%%%%%

% First pass check by Lotfi El Hafi on August 9th, 2019.
% Second pass check by Lotfi El Hafi on November 8th, 2019.

\subsubsection{Result of tidying up objects during planning phase}
\label{sub:experiment_result:planning}

\begin{figure}
    \centering
    \includegraphics[width=0.6\textwidth]{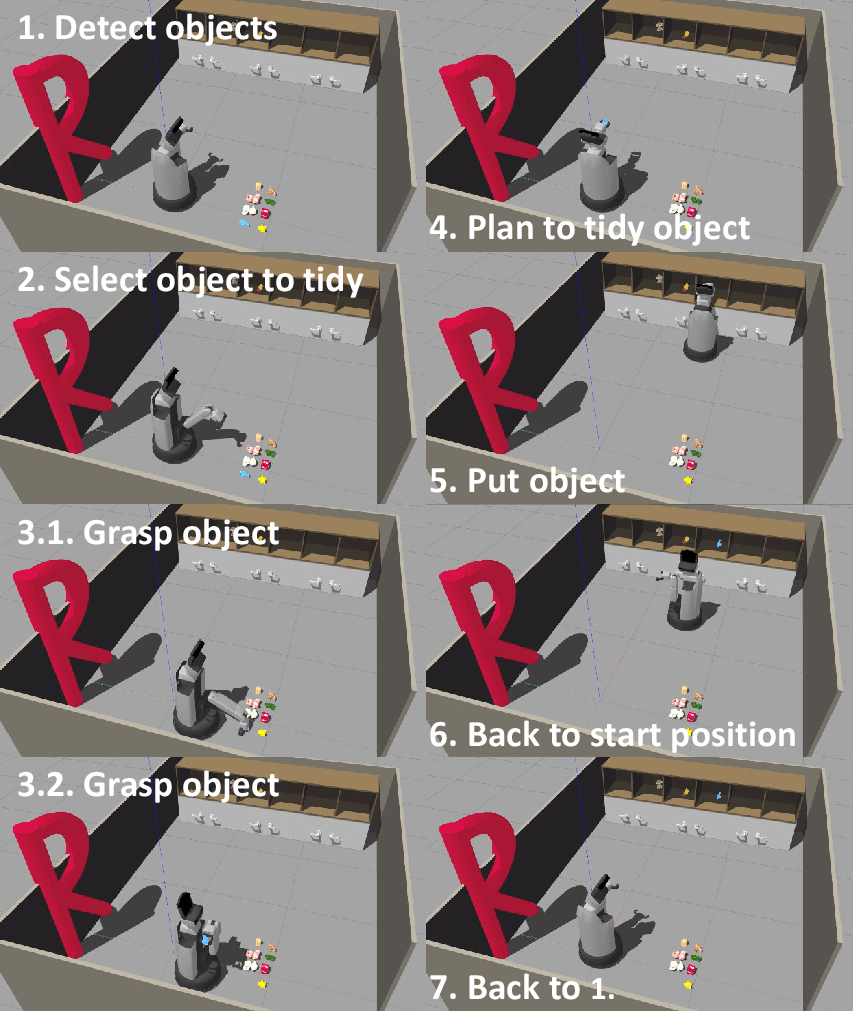}
    \caption{
        Process flow of tidying up objects performed by proposed method.
        Video example in which robot tidies up objects in Stage 1 can be found online at: {\url{https://youtu.be/inm1FHclibw}}.
    }
    \label{fig:tidy_flow}
\end{figure}

\begin{figure}
    \centering
    \includegraphics[width=0.58\textwidth]{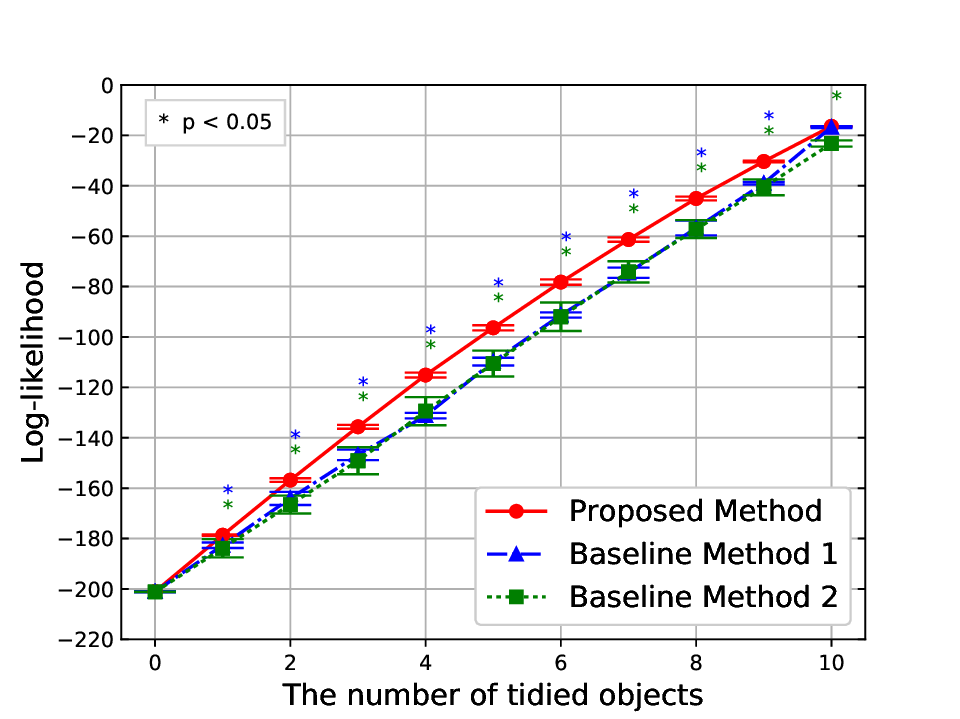}
    \caption{
        Log-likelihood values for each tidied object (Stage~1).
    }
    \label{fig:likelihood_stage1}
\end{figure}

\begin{figure}
    \centering
    \includegraphics[width=0.58\textwidth]{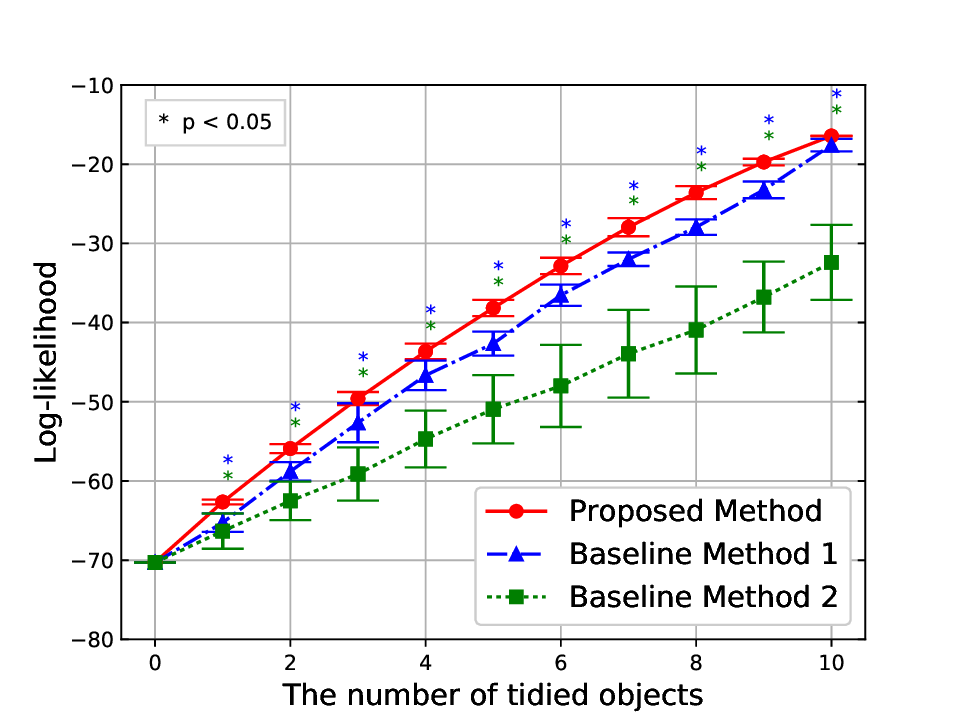}
    \caption{
        Log-likelihood values for each tidied object (Stage 2-1).
    }
    \label{fig:likelihood_stage2-1}
\end{figure}

\begin{figure}
    \centering
    \includegraphics[width=0.58\textwidth]{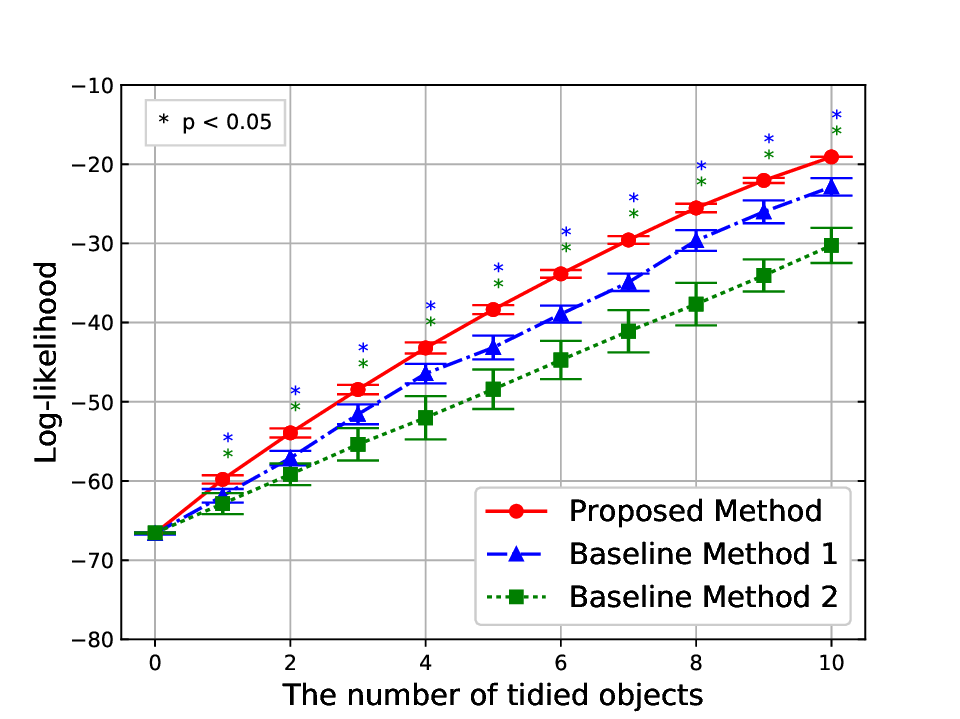}
    \caption{
        Log-likelihood values for each tidied object (Stage 2-2).
    }
    \label{fig:likelihood_stage2-2}
\end{figure}

Fig.~\ref{fig:tidy_flow} shows an example of the tidy-up process flow by the robot in Stage 1.
The robot operates for the number of objects determined in advance by repeating steps 1. -- 7. sequentially.
Even if grasping and release fail during the process, the robot can continue to operate continuously.

Figs.~\ref{fig:likelihood_stage1} -- \ref{fig:likelihood_stage2-2} show the average values and standard deviations of the log-likelihood values of the spatial concept model for each tidy-up planning in 10 trials of Stages 1, 2-1, and 2-2, respectively. 
In all the stages, the proposed method returned the highest values at any time.
Welch's t-test was performed to confirm that the log-likelihoods between the proposed method and each baseline are statistically significant.
The t-test showed that the changes for each step were statistically significant, except for the last step of baseline 1 in Fig.~\ref{fig:likelihood_stage1}.
Additionally, the t-tests on the mean of the log-likelihoods across the steps showed statistical significance in all the results.
Thus, the proposed planning method based on the spatial concept model is proven to be more efficient and accurate for selecting the objects to tidy up.

Table~\ref{tab:experiment_score_result_grasput} lists the score results for each method in Stages 1 and 2 in which the proposed method achieved the highest scores for both.
Because baseline 1 operated by taking the nearest object, the robot repeatedly failed to grasp the same nearest object, and the score was lower than that of the proposed method for both Stages 1 and 2.
In Stage~1, we considered that baselines 1 and 2 showed similar score values because the robot could achieve a score by simply moving an object anywhere in the toy boxes, even if the desired tidied position was incorrect.
In Stage 2-1, the score of baseline 2 was significantly lower because no point was obtained if the desired tidied position was incorrect.
Additionally, because Stage 2-2 contained objects with unknown tidied positions, the difference between the scores of the proposed method and baseline 1 was even more significant compared to that of Stage 2-1.
Because the robot could estimate the target tidied position of unknown objects by directly asking the user, the proposed method obtained a score higher than all the other methods in Stage 2-2.

Table~\ref{tab:experiment_score_result_no_grasput} shows the score of the planning without object grasping and release in 10 trials.
{Table~\ref{tab:experiment_accuracy_result_no_grasput} lists the accuracy rates of the methods for tidying up an object to the correct place.}
The proposed method showed again the highest values for both Stages 1 and 2.
In baseline 1, the position where to tidy up an object was selected from the list of the training data of the spatial concept model.
Therefore, we considered that it was easily influenced by the object position and the detection errors during the data acquisition.
We also considered that the proposed method led to a higher values because it was less likely to be affected by the data uncertainty when generalizing the target places to tidy up from the positions of the objects observed multiple times in the training data.
Additionally, the proposed method achieved a much higher value than the baselines in Stage 2-2 when objects with unknown tidied positions were included, as seen in Table~\ref{tab:experiment_score_result_grasput}.

\begin{figure}
    \centering
    \includegraphics[width=0.58\textwidth]{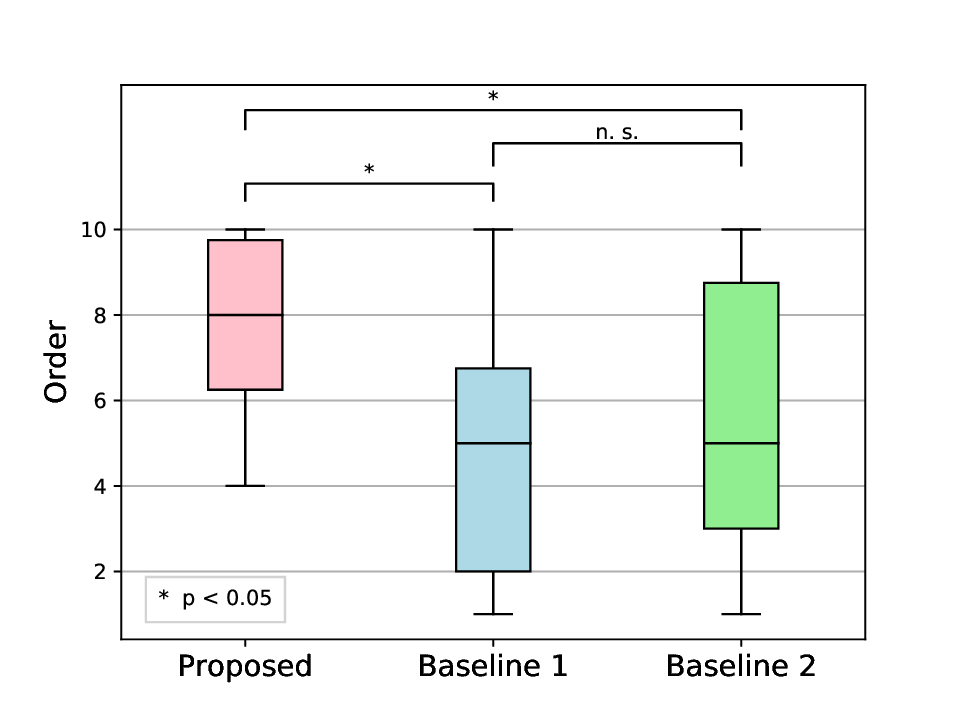}
    \caption{
        {Order in which unknown objects were tidied up in Stage~2-2.
        The statistical significance between the methods was checked by Mann-Whitney's U-test. A p-value of less than 0.05 was considered statistically significant.}
    }
    \label{fig:experiment_order_of_unknown}
\end{figure}

{
Fig.~\ref{fig:experiment_order_of_unknown} shows the order in which the unknown objects were tidied up in Stage~2-2.
For an object with an undefined or ambiguous destination place, the robot postponed tidying it up in the proposed method.
This means that the robot can tidy up objects with known destination places before asking the users.
This result lessens the burden on the users and allows for more efficient human-robot collaboration in tidy-up tasks.
}

\begin{table}
    \tbl{
        Score values {(rates)} of Tidy Up Here for each stage.
    }{
    \begin{tabular}{lrrr}
        \hline\noalign{\smallskip}
        \textbf{Method} & \textbf{Stage~1} & \textbf{Stage~2-1} & \textbf{Stage~2-2}\\
        \noalign{\smallskip}\hline\noalign{\smallskip}
        (1)~Proposed  & {\underline{\textbf{35}}}\hspace{2mm}{/50} {(0.70)} & {\underline{\textbf{28}}}\hspace{2mm}{/50} {(0.56)} & {\underline{\textbf{23}}}\hspace{2mm}{/59} {(0.39)}\\
        (2)~Baseline 1 & 14\hspace{2mm}{/50} {(0.28)} & 20\hspace{2mm}{/50} {(0.40)} & 10\hspace{2mm}{/59} {(0.17)}\\
        (3)~Baseline 2 & 13\hspace{2mm}{/50} {(0.26)} & 8\hspace{2mm}{/50} {(0.16)} & 4\hspace{2mm}{/59} {(0.07)}\\
        \noalign{\smallskip}\hline
    \end{tabular}
    }
    \label{tab:experiment_score_result_grasput}
\end{table}

\begin{table}
    \tbl{
        Score values {(rates)} of Tidy Up Here (without object grasping and release).
    }{
    \begin{tabular}{lrrr}
        \hline\noalign{\smallskip}
        \textbf{Method} & \textbf{Stage~1} & \textbf{Stage~2-1} & \textbf{Stage~2-2}\\
        \noalign{\smallskip}\hline\noalign{\smallskip}
        (1)~Proposed  & {\underline{\textbf{50}}}\hspace{2mm}{/50} {(1.00)} & {\underline{\textbf{50}}}\hspace{2mm}{/50} {(1.00)} & {\underline{\textbf{59}}}\hspace{2mm}{/59} {(1.00)}\\
        (2)~Baseline 1 & 49\hspace{2mm}{/50} {(0.98)} & 47\hspace{2mm}{/50} {(0.94)} & 40\hspace{2mm}{/59} {(0.68)}\\
        (3)~Baseline 2 & 34\hspace{2mm}{/50} {(0.68)} & 14\hspace{2mm}{/50} {(0.28)} & 16\hspace{2mm}{/59} {(0.27)}\\
        \noalign{\smallskip}\hline
    \end{tabular}
    }
    \label{tab:experiment_score_result_no_grasput}
\end{table}

\begin{table}
    \tbl{
        {Accuracy rate to tidy up an object to the correct place (with tidy-up planning only). The numerical values shows the mean and standard deviation in 10 trials.}
    }{
    {{
    \begin{tabular}{lllll}
        \hline\noalign{\smallskip}
        \textbf{Method} & \textbf{Stage~1} & \textbf{Stage~2-1} & \textbf{Stage~2-2} & \begin{tabular}{c} \textbf{Stage~2-2}\\(Unknown objects)\end{tabular}\\ %\shortstack
        \noalign{\smallskip}\hline\noalign{\smallskip}
        (1)~Proposed  & {\underline{\textbf{1.00}}} $\pm$ 0.000 & {\underline{\textbf{1.00}}} $\pm$ 0.000 & {\underline{\textbf{1.00}}} $\pm$ 0.000 & \hspace{6mm}{\underline{\textbf{1.00}}} $\pm$ 0.000 \\
        (2)~Baseline 1 & 0.97 $\pm$ 0.048 & 0.94 $\pm$ 0.070 {*} & 0.73 $\pm$ 0.125 {*} & \hspace{6mm}0.40 $\pm$ 0.263 {*}\\
        (3)~Baseline 2 & 0.22 $\pm$ 0.114 {*} & 0.27 $\pm$ 0.116 {*} & 0.25 $\pm$ 0.172 {*} & \hspace{6mm}0.33 $\pm$ 0.351 {*}\\
        \noalign{\smallskip}\hline\noalign{\smallskip}
        \multicolumn{5}{p{110mm}}{* Significant at 0.05 level in comparison between the proposed method and each baseline method.}
    \end{tabular}
    }}
    }
    \label{tab:experiment_accuracy_result_no_grasput}
\end{table}

%%%%%%%%%%%%%%%%%%%%%%%%%%%%%%%%%%%%%%%%%%%%%%%%%%%%%%%%%%%%%%%%%%%%%%%%%%%%%%%%

\section{Discussion}
\label{sec:discussion}
In this section, we discuss the applicability and limitations of our proposed method, as well as the avenues for future work.

{
First, we highlight the usefulness of adopting the WRS scoring system in our study.
The WRS scoring system is an open benchmark being sought as an evaluation method in the academic community of service robotics because it considers gradual levels of technical difficulties to assess the performance given a general service task.
With such an evaluation method, different robotics solutions can be objectively compared with criteria related to the task goal itself.
Although conventional metrics, such as time or accuracy, are appropriate for narrow or expert tasks, they can also be subject to many interpretations or assumptions when used for evaluating general service tasks that may not be as clearly delimited.
Therefore, by using an objectively comparable scoring system, we can evaluate the global performance of our proposed solution given a general tidy-up task, which ultimately is more important for the user.
We hope that studies conducted in the future will be evaluated using similar benchmarks.
In addition, our evaluation only used a part of the WRS scoring system but we would like to extend future evaluations to the full scoring system.
}

{
Next, we recognize that only a simulation space was used in the experiments of this study.
However, our simulation environment allowed to randomly select objects and furniture, and modify their positions. 
In future work, we plan to implement and evaluate the proposed method in various simulation environments and real-world spaces, but this implies further investigations.
For example, in a real-world environment, there is not always a unique destination to tidied an object.
The case of multiple destinations was not covered in the experiments in this paper, but the proposed method should be able to deal with such uncertainty.
For instance, when the likelihood is high for multiple destinations, the place with the highest likelihood will be set as the destination, i.e, the place where the robot has most observed the object.
}

{
The proposed method is also easily extendable by adding various conditions to the objective function of the tidy-up planning, e.g., the cost of the movement from the robot to the object.
In addition, if the user wants the robot to ask about the unknown objects first, it is possible to allow it by adding a negative term to the likelihood.
}
Moreover, although we conducted virtual simulations in this study, the proposed method is by nature environment-agnostic and can also work effectively in real environments.
Therefore, we consider the applicable scope of the proposed method to be broad.
In particular, the transfer of knowledge from a virtual to real environment would also be possible when the spatial concepts are learned in a simulator environment that imitates the real one, as illustrated by Stage 2 and shown in Fig.~\ref{fig:env_stage2}.
These two ideas will be investigated as future studies.

Another concern is that the one-hot vector of the object recognition results detected by YOLO was used as object information in the experiment.
In the future, we plan to test more complex tidy-up planning tasks considering environments where different objects classified into the same class exist in the recognition results at multiple separate places to tidy up.
Finally, the proposed method can also use the image features extracted by a CNN in the detected object area~\cite{ref:hatori2018interactively}.
In such a case, additional computational resources and time are required to calculate the object features; however, we believe these allow autonomously determining the positions where to move the objects based on the similarity of the features even for unknown new objects.

%%%%%%%%%%%%%%%%%%%%%%%%%%%%%%%%%%%%%%%%%%%%%%%%%%%%%%%%%%%%%%%%%%%%%%%%%%%%%%%%

% First pass check by Lotfi El Hafi on July 25th, 2019.
% Second pass check by Lotfi El Hafi on November 8th, 2019.

\section{Conclusion}
\label{sec:conclusion}

We proposed an autonomous tidy-up planning method based on a PGM for spatial concept formation using multimodal observations.
We also developed an autonomous robotic system that can tidy up home environments with several scattered objects.
The proposed method can learn co-occurrence probability between the objects and the target places where the objects should be tidied up from observation information collected by the robot.
Additionally, the robot can plan the tidy-up in order, starting from the object whose target place is the most defined among the objects cluttered in the environment using the parameters of the learned model.
In the experiment, we reproduced the Tidy Up Here task of the WRS international robotics competition using simulation environments constructed with Gazebo.
The results showed that the proposed method achieves efficient and accurate planning to tidy up scattered objects compared to other baseline methods.
Therefore, we consider that the proposed method is effective to tidy up various indoor environments, as reproduced by the WRS tasks.

%%%%%%%%%%%%%%%%%%%%%%%%%%%%%%%%%%%%%%%%%%%%%%%%%%%%%%%%%%%%%%%%%%%%%%%%%%%%%%%%

\section*{Acknowledgments}
This study was partially supported by the Japan Science and Technology Agency (JST) Core Research for Evolutionary Science and Technology (CREST) research program, under Grant JPMJCR15E3, and by the Japan Society for the Promotion of Science (JSPS) KAKENHI, under Grant JP20K19900.

%%%%%%%%%%%%%%%%%%%%%%%%%%%%%%%%%%%%%%%%%%%%%%%%%%%%%%%%%%%%%%%%%%%%%%%%%%%%%%%%

%\section*{Conflict of interest}

%The authors declare that they have no conflict of interest.

%%%%%%%%%%%%%%%%%%%%%%%%%%%%%%%%%%%%%%%%%%%%%%%%%%%%%%%%%%%%%%%%%%%%%%%%%%%%%%%%

% TODO: Replace '{\_}' by '_' in the URL fields inside 'jint.bib' if any.
\bibliographystyle{tADR}
\bibliography{ar}

%%%%%%%%%%%%%%%%%%%%%%%%%%%%%%%%%%%%%%%%%%%%%%%%%%%%%%%%%%%%%%%%%%%%%%%%%%%%%%%%

%\begin{comment}
\clearpage
\appendix
\section{{WRS Partner Robot Challenge: Tidy Up Here task evaluation}} 
\label{apdx:tidy-up}

We give the definition of Tidy Up Here, and of its Stages 1 and 2. 

%%%%%%%%%%%%%%%%%%%%%%%%%%%%%%%%%%%%%%%%

% First pass check by Lotfi El Hafi on August 2nd, 2019.
% Second pass check by Lotfi El Hafi on November 6th, 2019.

%Hence pick-and-place
Tidy up tasks with many kinds of objects in home environments are becoming a common problem in robotics applications, and competitions, e.g., the Tidy Up My Room Challenge, are held to invite the community to develop novel solutions.
In the Partner Robot Challenge (Real Space) of WRS, the Tidy Up Here task provides a benchmark for robot assistance, not only for disabled people but also for elderly people and healthy people, by supporting daily housekeeping tasks.
Therefore, tidy-up tasks are seen as an important issue to be addressed for the improvement of social welfare as well as the development of robotics in general.

%%%%%%%%%%%%%%%%%%%%%%%%%%%%%%%%%%%%%%%%

% First pass check by Lotfi El Hafi on August 2nd, 2019.
% Second pass check by Lotfi El Hafi on November 6th, 2019.

\begin{figure}
    \centering
    \includegraphics[width=0.70\textwidth]{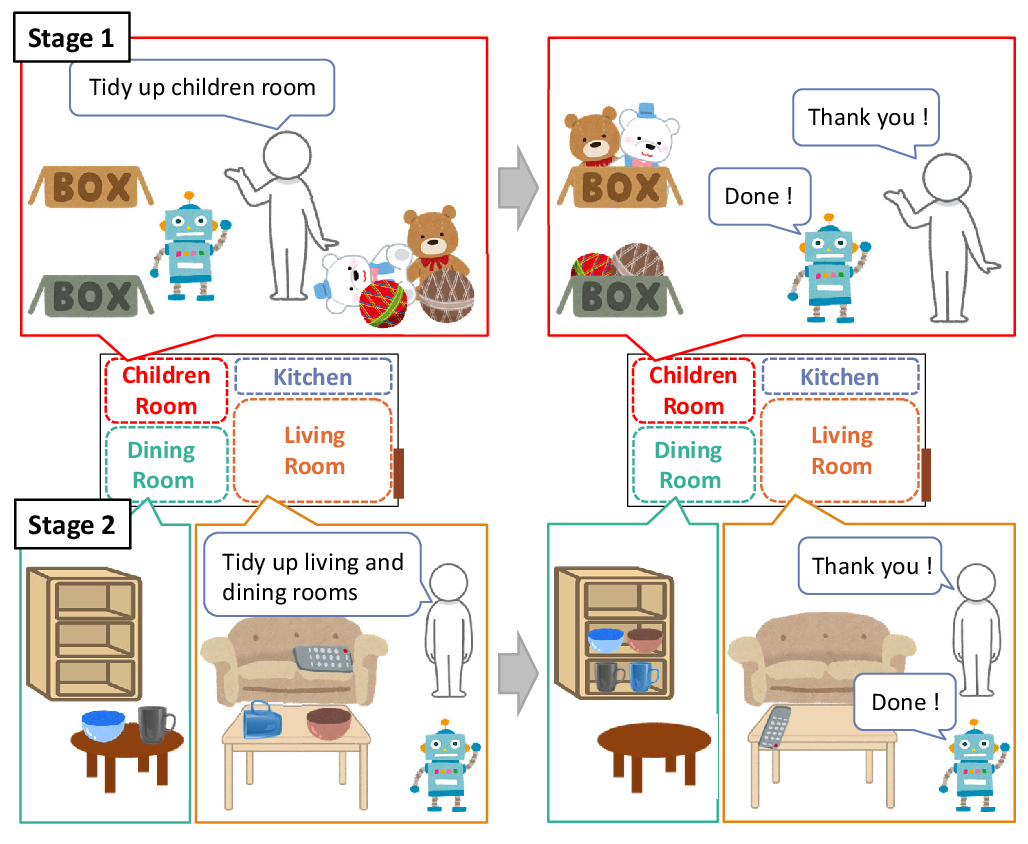}
    \caption{
        Overview of Tidy Up Here task that was conducted as one of tasks of Partner Robot Challenge (Real Space) at WRS 2018.
        Top: Stage 1, Bottom: Stage 2.
    }
    \label{fig:overview_tidyuphere}
\end{figure}

The WRS Tidy Up Here task consists of moving objects from incorrect positions to predetermined spaces, i.e. their original positions in the room, to consider them as tidied up.
It includes two separate sub-tasks of increasing challenge, designated as Stages 1 and 2 (See Fig.~\ref{fig:overview_tidyuphere}).
Regarding the environment, the layout of the room furniture is announced in advance and contains four rooms, e.g., a children room, a dining room, a kitchen, and a living room, with the names of the rooms known in the environment.
Additionally, there are two types of objects: known and unknown.
Known objects are provided with pre-announced recognition data and consist of approximately 45 units, without deformable objects, of frequently used daily goods and toys with information about their desired tidy up spaces, e.g., toys are on toy storage, foods are in refrigerator.
Unknown objects are not provided with pre-announced recognition data and place information.
They consist of approximately 10 units of daily items, including deformable objects, e.g., clothing, food, and paper.
For unknown objects, the robot can, from its own judgment, inquire the operator to provide the information on the desired tidy up space by voice or QR code, e.g., `Where should I put this object up?'.

Each stage starts with an instruction by the operator in the kitchen.
The robot then enters the rooms in a messy condition, and tidy up each stage in 12 minutes per trial.
The number of trials is two for each stage.

As described above, various information on the objects and places are provided beforehand to the participants.
However, in reality, it is necessary for the robot to acquire additional information about the real world from the object and place recognition to translate them into a form that can be understood and used by the robot.
Consequently, participating teams are provided a setup time to create/adjust the map data in the actual environment, and memorize unknown objects if necessary.

%%%%%%%%%%%%%%%%%%%%

% First pass check by Lotfi El Hafi on August 2nd, 2019.
% Second pass check by Lotfi El Hafi on November 6th, 2019.

\subsection{Tidy Up Here: Stage 1}
\label{apdx:tidy-up:stage1}

\begin{table}
    \tbl{
        Score table of Tidy Up Here at WRS (Stage 1).
    }
    {
    \begin{tabular}{p{100mm}r}
        \hline\noalign{\smallskip}
        \textbf{Performance} & \textbf{Score}\\
        \noalign{\smallskip}\hline\noalign{\smallskip}
        1. Tidy up an object into the toy storage & $3\times10$\\ 
        2. Tidy up an object into the correct box within the toy storage & $2\times10$\\ 
        3. Report back to the kitchen within the time limit & $2$\\
        \noalign{\smallskip}\hline\noalign{\smallskip}
        \textbf{Special Bonus} &\\
        \noalign{\smallskip}\hline\noalign{\smallskip}
        1. Complete the task on time & $3\times$ minutes remaining\\
        2. Open the house door & $20$\\
        \noalign{\smallskip}\hline
    \end{tabular}
    }
    \label{tab:tidyup_stage1_score}
\end{table}

Stage 1 consists of tidying objects scattered on the floor after a child has played.
The task starts with the robot being instructed to `Tidy up the children room.' by the operator.
In this simulated children room, there are 10 objects randomly selected from 15 known objects included in 5 categories, each of which containing 3 units.
The storage places are toy boxes.

Table~\ref{tab:tidyup_stage1_score} summarizes the evaluation criteria and the score table of Stage 1.
The robot gets 3 points for each object moved to the toy box.
If the object is placed in the correct box, the robot gets two additional points.
Additionally, the robot can get two points if it finishes the tidy-up task within the time limit and reports to the operator.
The objects scattered on the floor may become problematic obstacles when the robot moves.
Therefore, Stage 1 evaluates the basic and comprehensive ability of the autonomous tidy-up system of the robot.

%%%%%%%%%%%%%%%%%%%%

% First pass check by Lotfi El Hafi on August 2nd, 2019.
% Second pass check by Lotfi El Hafi on November 6th, 2019.

\subsection{Tidy Up Here: Stage 2}
\label{apdx:tidy-up:stage2}

\begin{table}
    \tbl{
        Score table of Tidy Up Here at WRS (Stage 2).
    }{
    \begin{tabular}{p{100mm}r}
        \hline\noalign{\smallskip}
        \textbf{Performance} & \textbf{Score}\\
        \noalign{\smallskip}\hline\noalign{\smallskip}
        1. Tidy up an object to the correct place & $5\times10$\\
        2. Inquire the human operator whether to discard or the tidy up space of the objects & $6\times5$\\
        3. Tidy up an unknown undeformable object & $3\times3$\\
        4. Tidy up an unknown deformable object & $3\times2$\\
        5. Report back to the kitchen within the time limit & $3$\\
        \noalign{\smallskip}\hline\noalign{\smallskip}
        \textbf{Special Bonus} &\\
        \noalign{\smallskip}\hline\noalign{\smallskip}
        1. Complete the task on time & $3\times$ minutes remaining\\
        2. Open the cabinet drawer & $20$\\
        3. Open the refrigerator door & $20$\\
        \noalign{\smallskip}\hline
    \end{tabular}
    }
    \label{tab:tidyup_stage2_score}
\end{table}

Stage 2 reproduces the conditions found when tidying a messy home environment.
The task starts with the robot being instructed to `Tidy up the living and dining rooms.' by the operator.
In Stage 2, 10 objects, consisting of 5 known objects and 5 unknown objects, are randomly selected from 32 known objects and 8 unknown objects.
The storage places are a coffee table, a wall shelf, a food cabinet, a kitchen unit, and so on.
The robot can ask the operator where is the desired tidy-up space.

Table~\ref{tab:tidyup_stage2_score} lists the evaluation criteria and the score table in Stage 2.
The robot gets 5 points for each object moved back to its correct original place.
If an unknown object is moved to the correct place, the robot gets 3 additional points.
Additionally, the robot can obtain 3 points if it finishes the tidy-up task within the time limit and reports to the operator.
Compared to Stage 1, the robot needs to deal with a larger range of objects and places.
It is difficult to finish the searching, moving, and tidying up of all objects within the time limit.
Therefore, the task conditions in Stage 2 are more advanced and difficult than in Stage 1.

%%%%%%%%%%%%%%%%%%%%

% First pass check by Lotfi El Hafi on August 8th, 2019.
% Second pass check by Lotfi El Hafi on November 7th, 2019.

\section{Detailed formulation of the Gibbs sampling}
\label{apdx:proposed_method:learning:gibbs}

We describe the sampling procedure of each parameter by Gibbs sampling as follows:
\begin{eqnarray}
    C_{i} &\sim& p\left( C_{i}=k \mid x_{i}, \mu, \Sigma, \pi, \varphi, \eta \right) \notag\\
    &&\propto \mathcal{N}\left( x_{i}\mid\mu _{C_{i}}, \Sigma _{C_{i}} \right) {\rm Mult}\left( o_{i} \mid \varphi _{C_{i}} \right) \notag\\
    &&\quad \times {\rm Mult}\left( w_{i} \mid \eta _{C_{i}} \right) {\rm Mult}\left( C_{i} \mid \pi \right), \label{eq:gibbs_start}\\
    \mu _{k}, \Sigma _{k} &\sim& \prod_{i=1}^{I} \mathcal{N}\left( x_{i}\mid\mu _{C_{i}=k}, \Sigma _{C_{i}=k} \right)
    \mathcal{NIW}\left( \mu _{k}, \Sigma _{k} \mid \mu _{0}, \kappa _{0}, \psi _{0}, \nu _{0} \right) \notag\\
    && \propto \mathcal{NIW}\left( \mu _{k}, \Sigma _{k} \mid \mu'_{k}, \kappa'_{k}, \psi'_{k}, \nu'_{k} \right),\\
    \varphi _{k} &\sim& \prod_{i=1}^{I} {\rm Mult}\left( o_{i} \mid \varphi _{C_{i}=k} \right) {\rm Dir}\left( \varphi _{k} \mid \alpha \right) \notag\\
    && \propto {\rm Dir}\left( \varphi _{k} \mid \alpha'_{k} \right),\\
    \eta _{k} &\sim& \prod_{i=1}^{I} {\rm Mult}\left( w_{i} \mid \eta _{C_{i}=k} \right) {\rm Dir}\left( \eta _{k} \mid \beta \right) \notag\\
    && \propto {\rm Dir}\left( \eta _{k} \mid \beta'_{k} \right),\\
    \pi &\sim& \prod_{i=1}^{I} {\rm Mult}\left( C_{i} \mid \pi \right) {\rm Dir}\left( \pi \mid \gamma \right) \notag\\
    && \propto {\rm Dir}\left( \pi \mid \gamma' \right),
    \label{eq:gibbs_end}
\end{eqnarray}
where the Gaussian-inverse-Wishart distribution is denoted as $\mathcal{NIW}(\cdot )$.
The hyperparameters of the posterior distribution $\alpha'_{k}, \beta'_{k},  \gamma', \mu'_{k}, \kappa'_{k}, \psi'_{k}, \nu'_{k}$ are calculated by the conjugate distributions between prior and likelihood.

%%%%%%%%%%%%%%%%%%%%

% First pass check by Lotfi El Hafi on August 8th, 2019.
% Second pass check by Lotfi El Hafi on November 7th, 2019.

\section{Formulation for simultaneous estimation of $N$ objects and their positions for tidying up}
\label{apdx:proposed_method:planning_formulation_multi}

The Tidy Up Here task has a time limit.
Specifically, among the number of scattered objects in the environment, the number of objects to be tidied up is finite. 
In this case, the problem of simultaneously finding $N$ objects and positions to be tidied up from the total number $D$ of detected objects is as follows:
\begin{eqnarray}
    \mathcal{D},\{x_{d}^{\ast}\} 
    = \argmax _{\mathcal{D},\{x_{d}^{\prime}\}} L \left( \{ x_{j} \}_{j \notin \mathcal{D}}, \{ x_{d}^{\prime} \}_{d\in \mathcal{D}} \right) - L \left( \{ x_{j} \} \right), 
    \label{eq:likelihood_function_multi}
\end{eqnarray}
\begin{eqnarray}
    &&L \left( \{ x_{j} \}_{j\notin \mathcal{D}}, \{ x_{d}^{\prime} \}_{d\in \mathcal{D}} \right) \notag\\
    &&= p \left( \{ x_{j} \}_{j\notin \mathcal{D}} \mid \{ o_{j} \}_{j\notin \mathcal{D}}, \Theta \right) p \left( \{ x_{d}^{\prime} \}_{d\in \mathcal{D}} \mid \{ o_{d} \}_{d\in \mathcal{D}}, \Theta \right) \notag\\
    &&\quad \propto \prod_{j \notin \mathcal{D}} \sum_{C_{j}} p \left( x_{j} \mid \mu_{C_{j}}, \Sigma_{C_{j}} \right) p \left( o_{j} \mid \varphi_{C_{j}} \right) p \left( C_{j} \mid \pi \right) \notag\\
    &&\qquad \times \prod_{d \in \mathcal{D}} \sum_{C_{d}} p \left( x_{d}^{\prime} \mid \mu_{C_{d}}, \Sigma_{C_{d}} \right) p \left( o_{d} \mid \varphi_{C_{d}} \right) p \left( C_{d} \mid \pi \right) 
    \label{eq:likelihood_function_multi2}
\end{eqnarray}
where the set of indexes $d$ of the selected objects to be tidied is denoted as $\mathcal{D}$.
Note that $1 \le N \le D$.

Similarly, considering the case of tidying $N$ objects at a time from the initial situation, the changing terms of the two equations of the likelihoods of before and after tidying up are:
\begin{eqnarray}
    &&\argmax _{\mathcal{D},\{x_{d}^{\prime}\}} \prod_{d \in \mathcal{D}} \cfrac{\sum_{C_{d}} p( x_{d}^{\prime} \mid \mu_{C_{d}}, \Sigma_{C_{d}}) p( o_{d} \mid \varphi_{C_{d}}) p( C_{d} \mid \pi)}{\sum_{C_{d}} p( x_{d}^{ } \mid \mu_{C_{d}}, \Sigma_{C_{d}}) p( o_{d} \mid \varphi_{C_{d}}) p( C_{d} \mid \pi )} \nonumber\\
    && = \prod_{n=1}^{N} \argmax _{d_{n},x_{d_{n}}^{\prime}} R_{n}({x_{d_{n}}, x_{d_{n}}^{\prime}}). 
    \label{eq:ratio_probability_algorithm}
\end{eqnarray}
The tidy-up method consisting of estimating combinations of $N$ objects at once shown in this section is equivalent to tidying object sequentially using a greedy method $N$ times, as described in Section~\ref{sec:proposed_method:planning_formulation_one}, because the object positions $x_{j}$ are conditionally independent from each other.
Finally, the robot can perform the tidy-up task by recursively selecting the object $d$ to be tidied up, one at a time, i.e. sequentially calculating Equation~(\ref{eq:ratio_probability_kai}).

%%%%%%%%%%%%%%%%%%%%%%%%%%%%%%%%%%%%%%%%

{
\section{Details on the implementation of the planning phase}
\label{apdx:proposed_system:planning_phase}
}

{
We give here practical details on the implementation of the planning phase as a supplement to Section~\ref{sec:proposed_system:planning_phase} and Fig.~\ref{fig:overview_proposed_system}~(g) in particular.
In this study, the developed system uses the integrated MoveIt!~\cite{ref:sucan2013moveit} framework, which relies on the open motion planning library (OMPL)~\cite{ref:sucan2012the-open-motion-planning-library} for planning the motion of the robot arm when grasping an object.
The motion planning algorithm used by MoveIt! adopts RRT-Connect~\cite{ref:kuffner2000rrt}, which is an extension of rapidly-exploring random trees (RRT)~\cite{ref:lavalle1998rapidly}.
The system uses Octomap~\cite{ref:hornung2013octomap} as obstacle information so that motion planning can be performed safely without collision.
Octomap is generated using a 3D map information generated in advance by real-time appearance-based mapping (RTAB-Map)~\cite{ref:labbe2014online}, which is a vSLAM method.
Furthermore, the robot moves to appropriate search positions while performing self-localization based on its map and observations.
When the robot cannot directly observe the position to tidy up an object from its current self-position, Octomap is used to navigate toward that position.
}

{
Additionally, we developed the system using high-level behavior state-machines implemented with FlexBE~\cite{ref:schillinger2016flexbe} because it provides various functional modules to realize the tidy-up task.
Using FlexBE allows to reuse the code and states in various systems and scenarios, and thus, increases abstraction during the development.
A video example in which the FlexBE running in our autonomous robotic system can be found online at: \url{https://youtu.be/Oj-p0Z6dHM0}.
}

%%%%%%%%%%%%%%%%%%%%%%%%%%%%%%%%%%%%%%%%

% First pass check by Lotfi El Hafi on August 9th, 2019.
% Second pass check by Lotfi El Hafi on November 8th, 2019.

\section{Pre-training for object detection}
\label{apdx:experiments:pre-train}

To detect the objects used in the experiment, we prepared a training dataset using an RGB-D camera mounted on the HSR and trained a YOLOv3 model.
The image size used for the object detection was $640 \times 480$ pixels.
A total of 10,000 images containing multiple objects were generated by augmentation using 20 images of different poses for each object and 25 background images of Stage 1 and Stage 2 environments.
Annotation files, including bounding box information of the objects in the augmented images, were generated and divided into two datasets of 90\% training data and 10\% test data.
The training and parameter tuning of YOLOv3 were performed using convolutional layer weights\footnote{YOLO: \url{https://pjreddie.com/darknet/yolo/}} pre-trained with ImageNet~\cite{krizhevsky2012imagenet}.
Object classes were prepared individually for each object.

%%%%%%%%%%%%%%%%%%%%%%%%%%%%%%%%%%%%%%%%

% First pass check by Lotfi El Hafi on August 9th, 2019.
% Second pass check by Lotfi El Hafi on November 8th, 2019.

\section{Pre-evaluation of objects to tidy up}
\label{apdx:experiments:pre-evaluation}

We evaluate the achievement of the tidy-up task for the successful estimation of tidying objects and positions.
As a pre-evaluation, we evaluate the accuracy of object detection, grasping, and release through all of the trials in the experiments.
The YOLOv3 object detection model used in the experiments is trained using our created dataset, as described in Appendix~\ref{apdx:experiments:pre-train}.
Additionally, various object shapes are included in the experiments.
After considering the effects of these above factors, we examine the superiority of the proposed planning method in Section~\ref{sub:experiment_result}.
The accuracy of object detection is defined as the ratio of the number of bounding boxes in which the object class is correctly detected to the number of bounding boxes detected and the number of objects not detected in the image.
The accuracy of object grasping and release is defined as the ratio of the number of successful actions to the total number of planned actions.

Table~\ref{tab:experiment_assumption_evaluation} summarizes the resulted accuracy of object detection, grasping, and release.
The object detection was highly accurate.
Additionally, the accuracy values of the object grasping and release were approximately 70\% although objects of various shapes were used.
In the case of grasping failure, either the grasp position coordinates of the detected object were inaccurately estimated, or the object shapes were complex to grasp, e.g., cylinders or stars. 

\begin{table}
    \tbl{
        Results of pre-evaluation for tidying up tasks.
    }{
    \begin{tabular}{lccc}
        \hline\noalign{\smallskip}
        \textbf{Accuracy} & \textbf{Stage~1} & \textbf{Stage~2-1} & \textbf{Stage~2-2}\\
        \noalign{\smallskip}\hline\noalign{\smallskip}
        Object detection & 1.00 & 0.96 & 0.95\\
        Object grasping & 0.64 & 0.69 & 0.76\\
        Object release & 0.84 & 0.79 & 0.68\\
        \noalign{\smallskip}\hline
    \end{tabular}
    }
    \label{tab:experiment_assumption_evaluation}
\end{table}
%\end{comment}

%%%%%%%%%%%%%%%%%%%%%%%%%%%%%%%%%%%%%%%%%%%%%%%%%%%%%%%%%%%%%%%%%%%%%%%%%%%%%%%%

\end{document}